\documentclass[10pt,journal,compsoc]{IEEEtran}

\usepackage[ruled, vlined]{algorithm2e}
\usepackage{algpseudocode}
\usepackage{amsmath}
\usepackage{fontawesome}
\usepackage{diagbox}
\usepackage{graphics}
\usepackage{subfigure}
\usepackage{epsfig}
\usepackage{threeparttable}
\usepackage{xspace}
\usepackage{graphicx}
\usepackage{cancel}
\newcommand\Ccancel[2][black]{
    \let\OldcancelColor\CancelColor
    \renewcommand\CancelColor{\color{#1}}
    \cancel{#2}
    \renewcommand\CancelColor{\OldcancelColor}
}
\usepackage{threeparttable}
\usepackage{color}
\usepackage[normalem]{ulem}
\usepackage{multirow}
\usepackage{float}
\usepackage{amsfonts}
\usepackage{bm}
\usepackage{array}
\usepackage[table]{xcolor}
\usepackage{colortbl}
\usepackage{rotating}
\usepackage{booktabs}
\usepackage{overpic}
\usepackage{contour}
\usepackage{enumitem}
\usepackage{cite}
\usepackage{tcolorbox}
\usepackage{amssymb}
\usepackage{pifont}
\usepackage{ragged2e}

\usepackage{csquotes}
\tcbuselibrary{theorems}
\tcbset{highlight math/.append style={left=0mm,right=0mm,top=0mm,bottom=0mm, colframe=white}}

\usepackage{amssymb}
\usepackage{mathtools}
\usepackage{amsthm}

\theoremstyle{plain}

\definecolor{lightgray}{gray}{.9}
\definecolor{deepgray}{gray}{.8}

\definecolor{mygray}{gray}{.9}
\newcolumntype{I}{!{\vrule width 1pt}}

\makeatletter
\newcommand{\thickhline}{%
    \noalign {\ifnum 0=`}\fi \hrule height 1pt
    \futurelet \reserved@a \@xhline
}
\makeatother

\makeatletter
\DeclareRobustCommand\onedot{\futurelet\@let@token\@onedot}
\def\@onedot{\ifx\@let@token.\else.\null\fi\xspace}

\definecolor{mygray}{gray}{.9}
\definecolor{mygreen}{RGB}{93,173,85}
\definecolor{mywarning}{RGB}{233,144,61}

\newcommand{\pub}[1]{{\color{gray}{\small{[{#1}]}}}}

\usepackage[pagebackref,breaklinks,colorlinks]{hyperref}
\usepackage[capitalize]{cleveref}
\Crefname{table}{Table}{Tables}
\crefname{table}{Tab.}{Tabs.}
\crefname{section}{§}{§§}

\newcommand{\rgmps}{{RGMP-S}}
\newcommand{\gss}{{LGSS}}
\newcommand{\wqkv}{{RASNet}}
\newcommand{\spike}{{SDFE}}
\newcommand{\adspike}{{ASN}}

\usepackage{tikz,xcolor}
\definecolor{lime}{HTML}{A6CE39}
\DeclareRobustCommand{\orcidicon}{
\begin{tikzpicture}
\draw[lime, fill=lime] (0,0)
circle[radius=0.16]
node[white]{{\fontfamily{qag}\selectfont \tiny \.{I}D}};
\end{tikzpicture}
\hspace{-2.5mm}
}
\foreach \x in {A, ..., Z}{%
\expandafter\xdef\csname orcid\x\endcsname{\noexpand\href{https://orcid.org/\csname orcidauthor\x\endcsname}{\noexpand\orcidicon}}
}

\begin{document}

\title{Generalizable Geometric Prior and Recurrent Spiking Feature Learning for Humanoid Robot Manipulation}

\author{Xuetao Li\hspace{-1mm}\orcidA{}, Wenke Huang\hspace{-1mm}\orcidB{}, Mang Ye\hspace{-1mm}\orcidC{},~\IEEEmembership{Senior Member,~IEEE}, Jifeng Xuan\hspace{-1mm}\orcidD{},~\IEEEmembership{Member,~IEEE}\\ Bo Du \hspace{-1mm}\orcidG{},~\IEEEmembership{Senior Member,~IEEE}, Sheng Liu\hspace{-1mm}\orcidF{},~\IEEEmembership{Fellow,~IEEE}, and Miao Li\hspace{-1mm}\orcidE{},~\IEEEmembership{Senior Member,~IEEE}

\IEEEcompsocitemizethanks{
\IEEEcompsocthanksitem 
The authors are with the School of Computer Science, School of Robotics, Institute of Technological Sciences, Wuhan University, Wuhan, China 430072. \protect
E-mail:\{xtli312, wenkehuang, yemang, jxuan, dubo, shengliu, miao.li\}@whu.edu.cn, (Xuetao Li and Wenke Huang contributed equally, Jifeng Xuan and Miao Li are corresponding authors).

\IEEEcompsocthanksitem A preliminary version of this work has appeared in Proceedings of the AAAI Conference on Artificial Intelligence (AAAI 2026) {\cite{FCCL_CVPR22}}.

}
}

\markboth{Generalizable Geometric Prior and Recurrent Spiking Feature Learning for Humanoid Robot Manipulation}%
{Shell \MakeLowercase{\textit{et al.}}: Bare Demo of IEEEtran.cls for Journals}

\IEEEtitleabstractindextext{
\begin{abstract}
\justifying
Humanoid robot manipulation is a crucial research area for executing diverse human-level tasks, involving high-level semantic reasoning and low-level action generation. However, precise scene understanding and sample-efficient learning from human demonstrations remain critical challenges, severely hindering the applicability and generalizability of existing frameworks. 
This paper presents a novel \rgmps,  Recurrent Geometric-prior Multimodal Policy with Spiking features, facilitating both high-level skill reasoning and data-efficient motion synthesis. 
To ground high-level reasoning in physical reality, we leverage lightweight 2D geometric inductive biases to enable precise 3D scene understanding within the vision-language model. Specifically, we construct a Long-horizon Geometric Prior Skill Selector that effectively aligns the semantic instructions with spatial constraints, ultimately achieving robust generalization in unseen environments. 
For the data efficiency issue in robotic action generation, we introduce a Recursive Adaptive Spiking Network. We parameterize robot-object interactions via recursive spiking for spatiotemporal consistency, fully distilling long-horizon dynamic features while mitigating the overfitting issue in sparse demonstration scenarios. 
Extensive experiments are conducted across the Maniskill simulation benchmark and three heterogeneous real-world robotic systems, encompassing a custom-developed humanoid, a desktop manipulator, and a commercial robotic platform. Empirical results substantiate the superiority of our method over state-of-the-art baselines and validate the efficacy of the proposed modules in diverse generalization scenarios. To facilitate reproducibility, the source code and video demonstrations are publicly available at \url{https://github.com/xtli12/RGMP-S.git}.
\end{abstract}
}
\maketitle
\IEEEdisplaynontitleabstractindextext
\IEEEpeerreviewmaketitle
\newcommand{\tworow}[2]{
\begin{tabular}{@{}c@{}}
{#1}  \\
{(#2 $\! \%$)}
\end{tabular}
}

\newcommand{\simplecnn}{{SimpleCNN}}

\definecolor{black}{rgb}{0.0, 0.5, 1.0}
\definecolor{black}{rgb}{0.0, 0.0, 0.0}
\definecolor{majorblack}{rgb}{0.0, 0.0, 0.0}

\IEEEraisesectionheading{
\section{Introduction}\label{sec:introduction}}
\IEEEPARstart{H}{umanoid} robots represent a pivotal frontier in embodied intelligence, capable of merging high-level cognition with dexterous physical interaction~\cite{tong2024advancements,li2024okami}, largely propelled by the emergence of foundation models trained on massive-scale datasets~\cite{zitkovich2023rt}. However, in real-world deployment, robotic agents face inherently unstructured environments governed by complex physical laws, which differ significantly from the static internet data distributions used in pre-training \cite{intelligence2025pi05visionlanguageactionmodelopenworld, SPIDER_ICML25}. Due to the infinite variability of 3D scenes and the exorbitant cost of collecting high-fidelity interaction data, the prevailing "brute-force" scaling paradigm, which relies on exhaustive statistical memorization, becomes prohibitive and sample-inefficient for mastering dynamic manipulation tasks \cite{abu2025development}. Driven by such real-world challenges, augmenting architectures with explicit geometric reasoning and spatial awareness has been increasingly explored, as it equips agents with structural inductive biases to decipher physical constraints effectively ~\cite{skubis2023humanoid}.

\begin{figure}[t]
  \centering
    \includegraphics[width=\linewidth]{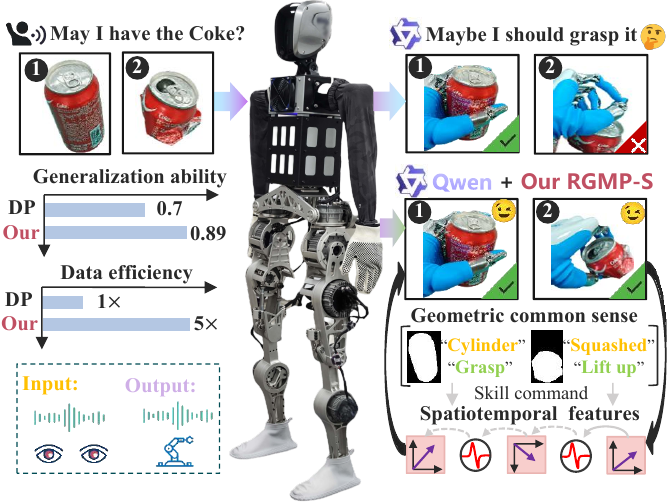}
  \vspace{-0.8cm}
   \caption{\textbf{Overview of our framework.} By integrating geometric commonsense  with spatiotemporal features, our \rgmps{} encodes the robot-target spatial relationships for manipulation tasks. \rgmps{} achieves a 19\% performance improvement and exhibits $5\times$ greater data efficiency compared to the Diffusion Policy (DP) baseline.}
     \vspace{-0.5cm}
   \label{fig:humanoid}
\end{figure}

Vision-Language Models (VLMs), exemplified by PaLM-E and InstructBLIP, excel at parsing spatial semantic information from multimodal inputs to generate logical robotic task plans~\cite{Driess2023PaLMEAE, dai2023instructblip, Lorasculpt_CVPR25}. However, a critical gap remains in the execution phase: current architectures struggle to translate abstract instructions into contextually appropriate robotic skills~\cite{team2024octo}. The fundamental bottleneck lies in the 3D geometric grounding process, where VLMs, despite their proficiency in object identification, often fail to infer the physical interaction affordances dictated by specific geometries~\cite{ahn2022can}. Specifically, regarding skill selection in open-world environments, this disconnect leads to a breakdown in differentiating manipulation primitives. Standard VLMs often confuse geometrically sensitive actions, such as grasping versus pinching, when encountering diverse object shapes \cite{shi2023robocook, tang2023graspgpt}. For instance, a model trained to grasp standard Coke cans typically targets the cylindrical sidewall. When presented with a crushed Coke can, the VLM retains semantic recognition but fails to perceive the structural deformation. Consequently, the robot attempts to execute the original side-grasping primitive, which is no longer physically feasible for the flattened geometry, resulting in task failure, as illustrated in \cref{fig:humanoid}. In fact, the flattened profile necessitates a pinching motion rather than the conventional side grasp. Such failures stem from a lack of explicit \textbf{geometric reasoning}. Existing approaches usually combine spatial indicators such as bounding boxes or contact points with high-level task descriptions \cite{mousavian20196, chu2025graspcot}. However, these methods predominantly rely on coarse semantic embeddings. In dynamic settings, such dependence results in plans that maintain semantic validity yet remain physically inexecutable \cite{rothert2024sim}. This deficiency emphasizes the critical necessity for systems capable of verifying skill feasibility through rigorous geometric analysis. This leads to a core research question: \textsl{\textbf{I)} How can robots effectively exploit geometric reasoning to reliably select feasible manipulation skills?}

Despite the remarkable progress in robotic manipulation driven by generative models, deploying these policies in dynamic environments remains a formidable challenge. A predominant paradigm, Diffusion Policies, has demonstrated exceptional capability in modeling action space by leveraging Gaussian noise to represent robotic behaviors \cite{chi2023diffusion}. By learning the gradient of the data distribution, these methods can generate high-fidelity trajectories that capture human-like dexterity \cite{pearce2023imitating}. However, this expressiveness comes at a steep computational cost. The inherent mechanism of diffusion models relies on an iterative reverse denoising process. As highlighted in recent analyses \cite{zitkovich2023rt, song2023consistency}, this sequential computation introduces prohibitive inference latency, rendering standard diffusion approaches ill-suited for high-frequency, responsive manipulation tasks where real-time feedback loops are critical. To circumvent the latency bottleneck of iterative solvers, recent advances in Vision-Language-Action (VLA) models and Transformer-based architectures advocate for direct, end-to-end action generation \cite{brohan2022rt, zitkovich2023rt, team2024octo}. These methods typically tokenize visual observations and directly predict actions in a feed-forward manner, ensuring rapid inference. However, achieving robust generalization with these architectures has historically demanded a brute-force strategy: scaling model parameters to billions and training on massive, diverse datasets. While effective at scale, this approach is data-inefficient and computationally heavy. Crucially, these methods often treat visual inputs as flattened sequences of semantic tokens, ignoring the intrinsic spatiotemporal mechanisms of manipulation. They lack explicit inductive biases to capture the fine-grained \textbf{spatial relationships} between the robotic end-effector and the target object, and often fail to distinguish task-relevant features from the dense, noisy visual context \cite{shridhar2022cliport, liang2023code}. Consequently, without massive data covering every variation, these black-box mappings are prone to latching onto spurious background correlations, leading to poor performance under domain shifts or in cluttered scenes \cite{NRCA_ICML25}. Thus, a critical research gap exists between the need for expressive, real-time control and the requirement for data-efficient generalization. The challenge lies in moving beyond simple parameter scaling or inefficient iterative denoising. Instead, we must explore how to construct an architecture that naturally embeds the spatiotemporal structure of manipulation tasks. This leads to the other core research question: \textsl{\textbf{II)} How can robots effectively learn robust data-efficient policies from limited demonstrations?}

To surmount the deficiency in spatial-geometric reasoning outlined in \textsl{\textbf{I)}}, inspired by the efficiency of human cognitive processes in synthesizing vision and logic, a preliminary version of this work published in AAAI 2026 \cite{FCCL_CVPR22} explores geometric priors via a lightweight commonsense tuning strategy. In particular, we aim to bridge the gap between 3D geometric perception and semantic task planning through implicit geometric information decoding from 2D inputs. Traditional explicit 3D reconstruction provides detailed spatial information, but it exhibits high computational latency and sensitivity to noise \cite{sitzmann2020implicit}, and lacks of adaptability for real-time robotic interaction. This motivates us to bypass costly reconstruction in favor of capturing implicit geometric structures directly from RGB inputs. We propose the \textbf{Long-horizon Geometric-prior Skill Selector} (\gss{}) to encode spatial affordances into the latent space of the policy, effectively aligning 3D spatial indicators with robotic action spaces and commonsense reasoning. Specifically, we integrate geometric inductive biases into a VLM with minimal fine-tuning, requiring only a concise set of rule-based constraints. This approach offers a practical solution to realize robust decision-making in unseen environments. Furthermore, we believe that \gss{} is essential because it introduces a decision-making process analogous to human cognition \cite{battaglia2018relational}, which mirrors the human ability to synthesize an action plan for the long-horizon manipulation task. {\color{black}
To this end, we develop the \gss{} framework to enable the manipulation of diverse objects via rigorous geometric consistency checks, proving its effectiveness and robustness in real-world humanoid deployment.}

\begin{figure*}[ht!]
  \includegraphics[width=\textwidth]{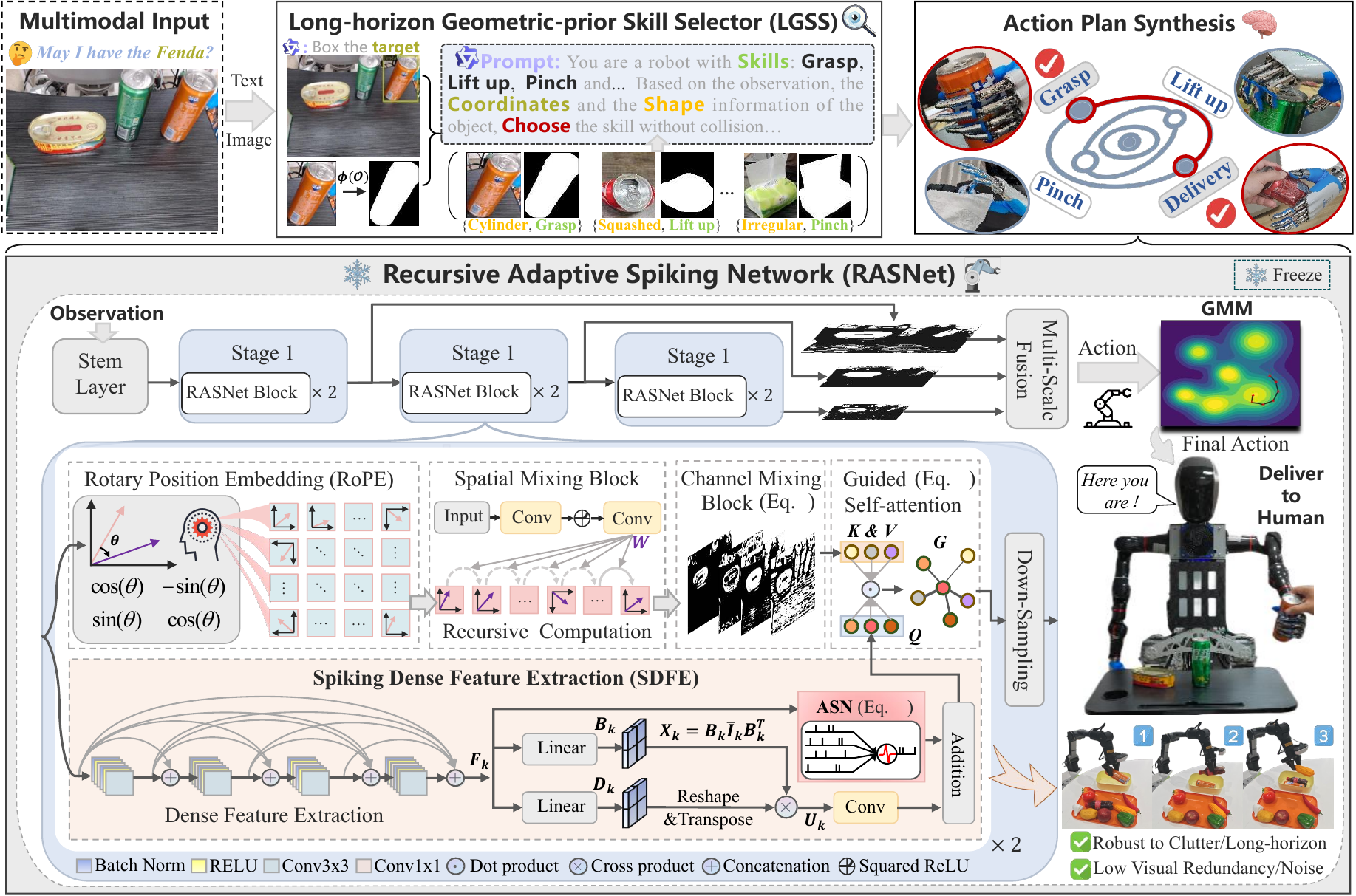}
  \put(-213.1,150.5){\makebox(0,0)[l]{\fontsize{7}{7}\selectfont\ref{eq:channel}}}
    \put(-155.5,159.5){\makebox(0,0)[l]{\fontsize{7}{7}\selectfont\ref{eq:GSNET}}}
        \put(-178.5,72.5){\makebox(0,0)[l]{\fontsize{7}{7}\selectfont\ref{eq:spike}}}
  \vspace{-0.4cm}
  \caption{\textbf{Pipeline of \rgmps{}}. Upon receiving a speech command, the robot utilizes \gss{} (see \cref{subsec:gss} for details) to identify and localize the target object. By integrating object coordinates, shape cues (extracted via Yolov8n-seg~\cite{yaseen2024yolov9} model \(\phi\)), and geometric-prior knowledge, the robot selects an appropriate skill from the library, where each primitive is associated with a pretrained \wqkv{} model  (see \cref{subsec:wqkv} for details). The selected \wqkv{} model subsequently executes the task precisely through adaptive recursive feature extraction and GMM-based refinement. }
\label{fig:pipeline}
\vspace{-0.4cm}
\end{figure*}

To surmount the overfitting risks inherent to limited data (\textsl{\textbf{II}}), we deviate from data-hungry scaling and instead embed explicit spatiotemporal inductive biases via recursive spiking dynamics, enabling the efficient distillation of robust skills from sparse demonstrations. In our conference version {\cite{FCCL_CVPR22}}, we employed recurrent computation to construct spatial relationships between the robotic end-effector and the target object, offering a foundational policy for action generation. However, standard recursive models are prone to vanishing gradients during global modeling \cite{pascanu2013difficulty} and often struggle to distinguish task-salient features from complex backgrounds \cite{bengio1994learning}. We posit that effective spatial memory requires dynamic regulation to prevent the loss of key historical cues while suppressing noise. Purely spatial processing lacks the temporal reasoning necessary to filter redundant information from dense visual features \cite{roy2019towards}. This motivates us to introduce an adaptive mechanism that integrates spiking information to refine feature retention. In this journal version, we propose the \textbf{Recursive Adaptive Spiking Network} (\wqkv{}), which incorporates Spiking Dense Feature Extraction (\spike{}) to embed temporal reasoning into spatial feature understanding. We utilize Rotary Position Embeddings (RoPE) and recursive computation to form a global spatial memory, while simultaneously deploying an Adaptive Decay Mechanism and Adaptive Spike Neurons (\adspike{}) to dynamically control memory retention and amplify task-critical features. Finally, we apply Gaussian Mixture Models (GMM) to approximate the joint motions. Compared with our conference version, \spike{} not only effectively suppresses redundancy within dense information but also significantly enhances the generalization ability in unseen scenarios featuring complex backgrounds.

Our primary contribution is to present \rgmps{}, a unified framework that bridges the semantic-geometric gap to achieve robust, data-efficient, and generalizable robotic manipulation. To successfully facilitate high-level skill reasoning and low-level motion synthesis, we present the following technical contributions:

\begin{itemize} \item \textbf{Long-horizon Geometric-prior Skill Selector (\gss{}).} We introduce a lightweight geometric tuning strategy that fuses vision-language models with low-rank geometric adapters. By injecting shape-level commonsense biases, \gss{} aligns semantic instructions with latent geometric constraints, enabling the precise selection of parameterized skills from a pretrained library without requiring task-specific fine-tuning.
\item \textbf{Recursive Adaptive Spiking Network (\wqkv{}).} We formulate a novel policy architecture to investigate intrinsic spatiotemporal mechanisms in robotic manipulation. By modulating latent representations through Adaptive Spike Neurons (\adspike{}) and recursive computation, this network captures directional spatial dependencies within a spatiotemporally consistent latent space. This design effectively distills task-critical features from dense visual observations while mitigating optimization conflicts in sparse data regimes.
\item \textbf{Data-Efficient Generalization.} We conduct a rigorous validation of the \rgmps{} framework across the ManiSkill benchmark and three diverse physical robotic platforms. Extensive real-world experiments demonstrate that our framework ensures robustness against dynamic disturbances, significantly outperforming the Diffusion Policy baseline with an 89\% generalization success rate and 5$\times$ superior data efficiency.
\end{itemize}

\noindent This manuscript represents a substantial extension of our preliminary version published in AAAI 2026 \cite{FCCL_CVPR22}. We augment the original work in the following aspects:
\begin{itemize}
\item \textbf{Algorithmic Advancement in Reasoning.} We advance the preliminary skill selector to the \gss{} via a hierarchical Chain-of-Thought (CoT) mechanism. This enhancement enables the structured decomposition of intricate goals into multi-stage sequences, significantly boosting performance on long-horizon manipulation tasks (e.g., towel folding and water pouring).

\item \textbf{Architectural Evolution for Robustness.} We evolve the previous visuomotor module with Spiking Dense Feature Extraction (\spike{}) and Adaptive Spike Neurons (\adspike{}) to dynamically filter redundant visual noise in cluttered scenes (as shown in \cref{fig:pipeline}). This addresses the limitations of the original model regarding feature redundancy and lack of temporal logic, ensuring robust action synthesis in unstructured environments.

\item \textbf{Expanded Real-World Evaluation.} We broaden the experimental scope by incorporating a commercial Aloha robot \cite{fu2024mobile} and introducing three additional challenging long-horizon manipulation tasks: towel folding, water pouring, and bin picking, as detailed in \cref{sec:ExperSetup}.
\end{itemize}

\section{Related Work}\label{sec:related}

\subsection{Vision-Language Models for Robotic Manipulation}
\noindent \textbf{Semantic Planning and Grounding}. A pioneering line of work leverages the semantic reasoning capabilities of Large Language Models (LLMs) and Vision-Language Models (VLMs) to decompose high-level instructions into a sequence of robotic primitives {\cite{ahn2022can, Driess2023PaLMEAE, dai2023instructblip}}. While these foundation models excel at high-level task planning, they often treat the physical world as a semantic abstraction. Consequently, they suffer from a \textit{semantic-geometric gap}, where the generated plans are logically sound but physically inexecutable due to a lack of low-level spatial awareness. Subsequently, a large body of methods {\cite{ TPAMI_VLM_Survey23, sun2025towards, liu2025diff9d, liu2025general}} 
 began to incorporate explicit 3D representations, such as voxels or neural radiance fields, to ground semantic queries into physical space. For instance, recent research has explored establishing dense correspondence between visual observations and linguistic concepts to enhance manipulation precision {\cite{qiu2021dense}}. These methods mainly focus on explicit 3D reconstruction or external trajectory solvers to bridge the gap. However, when robots operate in unstructured open-world environments, explicit reconstruction exhibits high computational latency and sensitivity to perception noise {\cite{sitzmann2020implicit}}. Meanwhile, relying on external motion planners separates semantic reasoning from control, which is a suboptimal setting for dynamic interaction. In this paper, we focus on learning implicit geometric priors directly from 2D RGB inputs. By fusing geometric adapters with VLMs, our \gss{} avoids costly reconstruction while maintaining robust spatial reasoning.

\noindent \textbf{Geometric Reasoning in Manipulation}.
With the demand for precise object interaction, geometric-aware manipulation has been an active research field. One direction is introducing object-centric representations {\cite{transporter_zeng2020, kpam_manuelli2019}} that predict keypoints or 6-DoF poses. However, these techniques struggle with category-agnostic objects or deformable materials where defining canonical poses is difficult. Recently, a specific line of research {\cite{tang2023graspgpt, shi2023robocook, TPAMI_Grasp_24}} leverages affordance maps or contact primitives to guide grasping. Therefore, these approaches heavily rely on the quality of pre-defined primitives and often fail to distinguish fine-grained manipulation modes (e.g., grasping versus pinching) under structural deformation. Recent works \cite{jin2024reasoning, zhao2024omdet, mousavian20196} have proven the feasibility of combining bounding boxes with language. However, these methods reach only coarse spatial alignment, which is not suitable for learning a generalizable policy for diverse geometries and thus lead to task failure in unseen scenarios. In this paper, based on the Geometric Prior Skill Selector, we inject shape-level commonsense biases to enforce rigorous geometric consistency checks, thereby effectively mitigating uncertainties and handling dynamic constraints to ensure robust reasoning in tasks.

\subsection{Generative Robotic Policy Learning}
Imitation learning {\cite{pomerleau1988alvinn, dagger_ross2011}} has been an essential paradigm in robotics, with the goal of acquiring robust policies from expert demonstrations \cite{razali2025keystate}. Existing works can be broadly divided into two branches: Transformer-based methods {\cite{act_zhao2023, brohan2022rt, decisiontransformer, hu2024transforming}} and Diffusion-based methods {\cite{chi2023diffusion, consistency_policy_2024}}. Transformer architectures, such as the Decision Transformer \cite{chen2021decision}, treat control as a sequence modeling problem. Specifically, self-attention mechanisms in Transformers have been shown to effectively capture global dependencies {\cite{TPAMI_Transformer_Robotics24}}. However, achieving generalization with Transformers has historically demanded a brute-force scaling strategy involving billions of parameters and massive datasets {\cite{zitkovich2023rt}}. In contrast, Diffusion Policies, they model the action distribution using iterative denoising processes. Despite their expressiveness in capturing multimodal distributions, the sequential denoising computation introduces prohibitive inference latency. Unlike standard diffusion approaches, we focus on a feed-forward recursive architecture that ensures high-frequency real-time responsiveness without the heavy computational strain typically associated with iterative solvers.

\subsection{Spatiotemporal Representation}
Learning effective spatiotemporal representations is critical for mastering dynamic tasks. Traditional approaches utilize Recurrent Neural Networks (RNNs) or LSTMs {\cite{lstm_hochreiter1997}} to model temporal dependencies. However, standard recursive models are prone to vanishing gradients and often struggle to filter redundant information from dense visual streams. Recently, Spiking Neural Networks (SNNs) have emerged as a powerful method for processing temporal information with high energy efficiency and sparse computation {\cite{ghosh2025event, wu2022brain, yao2025scaling}}. Related methods have explored SNNs for event-based vision and motion estimation {\cite{wu2022brain, TPAMI_Event_SNN23}}. The key difference between our \wqkv{} and the aforementioned efforts is that ours is designed for dense visual feature extraction rather than purely neuromorphic processing. Inspired by the efficiency of spiking neurons, we construct the Adaptive Spike Neuron (\adspike{}) to modulate feature retention. Furthermore, we investigate how to better maintain global spatial dependencies and develop the Recursive Adaptive Spiking Network, which captures intrinsic spatiotemporal mechanisms and ensures data efficiency under limited demonstrations.

\section{Methodology}\label{sec:Methodology}
\begin{table}[t]
\label{tab:notation}
\centering
\caption{
{\color{majorblack}
\small{
\textbf{Notations} table.
}
}
}
\vspace{-10pt}
\label{tab:notation}
\resizebox{\linewidth}{!}{
		\setlength\tabcolsep{2pt}
		\renewcommand\arraystretch{1.2}
\begin{tabular}{clIcl}
\hline\thickhline
\rowcolor{lightgray}
 & Description &  & Description\\ \hline\hline
$\mathcal{P}$ & Generated action plan &
$\mathcal{I}$ & Current user instruction \\
$\mathcal{O}$ & Current visual observation &
$\mathcal{C}$ & Predefined contextual set \\
$\mathcal{P}_i$ & Action sequence (\cref{eq:vlm}) &
$\mathcal{W}$ & Adaptive decay factors (\cref{eq:adm}) \\
$F_0$ & Initial feature map (\cref{eq:stem}) &
$X_k$ & Spatial attention mask (\cref{spacial}) \\
$B_k$ & Spatial attention feature (\cref{spacial}) &
$D_k$ & Feature for attention processing \\
$Q_A$ & Spiking dense feature (\cref{psnl}) &
$F_k$ & Extracted dense feature map \\
$\mathcal{A}$ & Adaptive spike neuron function &
$\mu_k$ & Mean of $k$-th GMM component \\
$F_f$ & Fused feature map (\cref{eq:GSNET}) &
$a_{in}$ & Initial action prediction (\cref{eq:ain}) \\
$\Sigma_k$ & Covariance of $k$-th GMM component &
$l_k$ & Mahalanobis distance (\cref{eq:regression}) \\
$K$ & Number of GMM components &
$\mu_k^\omega$ & Mean of $k$-th GMM component \\
\hline
\end{tabular}%
}
\vspace{-15pt}
\end{table}
\textbf{Overview  of \rgmps{}.}
To achieve generalizable robotic manipulation, we introduce \textbf{R}ecurrent \textbf{G}eometric-prior \textbf{M}ultimodal \textbf{P}olicy with \textbf{S}piking Features (\rgmps{}). This framework generates task plans and 6-DoF actions for humanoid robots using egocentric RGB inputs. \rgmps{} comprises two key components: the \textbf{L}ong-horizon \textbf{G}eometric-prior \textbf{S}kill \textbf{S}elector (\gss{}) and the \textbf{R}ecursive \textbf{A}daptive \textbf{S}piking \textbf{Net}work (\wqkv{}). The \gss{} translates natural language commands and visual observations into executable skill sequences by incorporating geometric commonsense priors. Complementarily, the \wqkv{} converts visual inputs into robotic manipulation actions through spatiotemporal memory modeling. 

Instead of employing computationally expensive explicit 3D reconstruction, the policy learns to extract implicit geometric information from RGB inputs by correlating spatial cues with action labels and commonsense reasoning. Additionally, we incorporate spiking neural networks to process dense features that capture fine-grained spatial perception. This approach enables the model to focus on task-relevant visual regions, emphasizing important details while suppressing redundant information, thus enhancing performance. This method is inspired by the natural and intuitive process of human object manipulation. In human grasping, the eyes first detect the target object, and the brain rapidly processes this sensory input to generate an appropriate motor command for the hand, allowing the arm to execute the grasping action and complete the task successfully. Similarly, our \rgmps{} mimics this process by seamlessly integrating perception with action: the RGB image is analyzed to identify the object and its relevant geometric context, while the \wqkv{} is employed to predict the joint angles required for the robotic arm to grasp the object. A summary of notations is provided in \cref{tab:notation}.

\subsection{Long-horizon Geometric-prior Skill Selector}
\label{subsec:gss}
\textbf{Motivation of Geometric Priors.} A fundamental challenge in robotics is the precise selection of skills (e.g., grasping versus pinching) for objects with diverse geometries or in previously unseen scenarios. While VLMs support object recognition and localization, they often struggle to translate semantic information into accurate actions, primarily due to their lack of explicit integration of \textbf{geometric priors} in the vision-action mapping process. This limitation motivates the development of our novel \gss{} framework, which seamlessly combines geometric reasoning with semantic task planning through geometric commonsense alignment.

The \gss{} operates through two sequential phases to bridge semantic instruction and physical execution. Initially, a VLM~\cite{bai2023qwen} is employed to interpret human instructions, allowing the robot to detect and locate the target object within the provided image. This visual grounding step generates bounding boxes that anchor subsequent reasoning processes. Building on this visual localization, the system then analyzes spatial relationships and morphological characteristics of the identified objects. This geometric analysis incorporates commonsense knowledge about object properties and their functional implications for manipulation. The resulting object-centric representation, combining semantic and geometric features, guides the selection of appropriate manipulation strategies from a predefined skill library. This structured approach enables the robot to translate abstract commands into context-aware action sequences through a novel geometric-object decomposition mechanism that coordinates perception with action selection:
\begin{equation}\small
\label{eq:vlm}
\mathcal{P} = plan(\mathcal{I},{\kern 1pt} {\kern 1pt} \mathcal{O}{\kern 2pt}|{\kern 2pt}\mathcal{C}\;),
\end{equation}
where $\mathcal{P}$ represents the generated action plan, $\mathcal{I}$ is the current natural language user instruction, $\mathcal{O}$ denotes the current visual observation, and $\mathcal{C}$ constitutes a predefined contextual set containing $n$ examples $\{({\mathcal{I}_i}, {\mathcal{O}_i}, {\mathcal{P}_i})\}_{i=1}^n$, which encapsulates instructions, prompts, and commonsense reasoning to facilitate in-context learning.

The system processes an RGB image \(\mathcal{O}\) where a Visual-Language Model first identifies the target object with a bounding box. This localization provides the foundation for generating executable manipulation skills. The model combines predefined context \(\mathcal{C}\) with geometric properties like spatial relationships and object shape to determine these skills. For example, when receiving the instruction \textit{``I want Fanta''}, the \gss{} follows the contextual prompt \textit{``Please box the target object in the instruction''}. It locates the specific beverage can among other items, then uses YOLOv8n-seg to determine its shape. Finally, the visual-language model integrates contextual information \(\mathcal{C}\) with geometric reasoning to produce operational commands.

The \gss{} adopts a modular design with minimal computational overhead. Its architecture integrates three key components: a visual-language interpretation module for command processing, a semantic segmentation module for object analysis, and a prompt-based skill selector for action determination. These elements operate synergistically to transform user instructions and visual inputs into appropriate skill choices. The complete inference process requires 105 milliseconds when executed on an NVIDIA 4090 GPU.

\noindent
\textbf{Visual-Language Interpretation Module:} This component employs the Qwen-vl API for visual-language understanding, utilizing structured prompt engineering to incorporate prior knowledge. The preparation involves providing the API with carefully selected training instances, each containing: (1) an RGB image with an object bounding box, (2) corresponding semantic segmentation results, (3) manually annotated shape categories (such as ``cylindrical'' or ``crushed''), and (4) the corresponding manipulation skill. For example, training data includes a Fanta can image with a bounding box and segmentation mask, labeled as ``cylindrical shape'' and paired with the ``side grasp''; similarly, an encircled Fanta can is associated with the ``lift up''. The model is primed with 20 representative examples spanning common object categories to establish a robust mapping between visual features, geometric attributes, and manipulation strategies. During operation, the API processes input prompts formatted as: \textit{``Instruction: Identify target object in image and output bounding box [x1, y1, x2, y2]''}. This configuration achieves 93.1\% localization accuracy on our dataset with a mean latency of 45 ms per query.

\noindent
\textbf{Semantic Segmentation Module:} This component processes the bounding box coordinates provided by the Qwen-vl API by extracting the corresponding image region and passing it to a YOLOv8-seg model. The model, which has been pre-trained on large-scale datasets and subsequently fine-tuned on our proprietary data, generates precise semantic segmentation masks to capture the object's geometric structure. On our evaluation dataset, the module achieves a mean Intersection over Union (mIoU) of 97.6\%, demonstrating the reliability of its shape extraction capabilities for subsequent geometric reasoning.

\begin{figure}[!t]
  \includegraphics[width=\columnwidth]{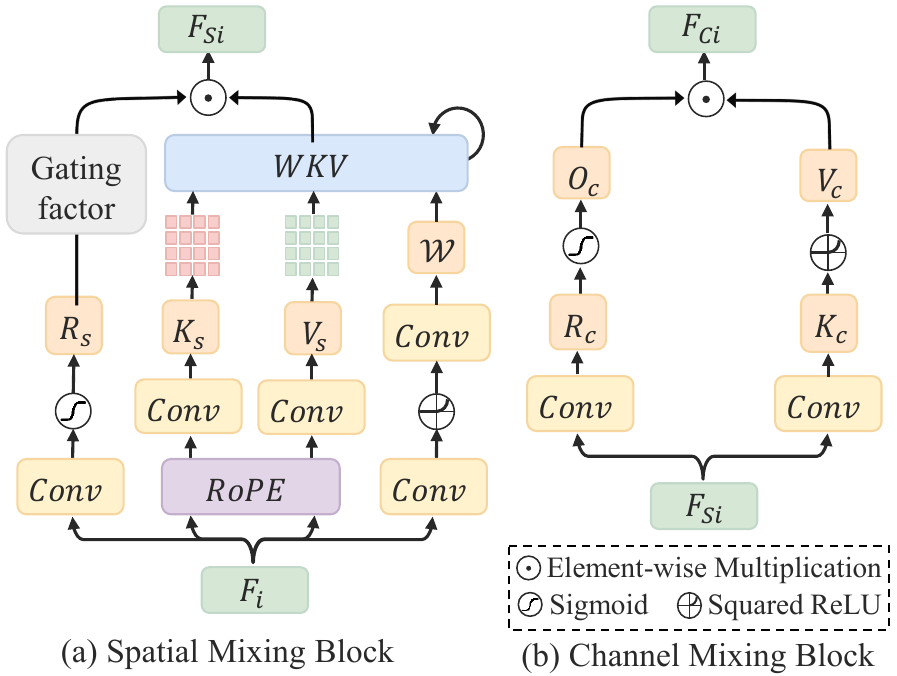}
    \vspace{-0.8cm}
  \caption{\textbf{Structure of Spatial Mixing Block and Channel Mixing Block.} The Spatial Mixing Block uses ADM to generate Dynamic Decay $\mathcal{W}$ recursively, and employs RoPE to introduce directional awareness relative to spatial positions. These mechanisms collectively enhance the ability of the block to integrate spatial information effectively. The Channel Mixing Block reallocates channel-wise feature responses by integrating correlations between channels. See \cref{subsec:wqkv} for details.}
\label{fig:RWKV}
  \vspace{-0.4cm}
\end{figure}

\noindent \textbf{Geometry-Grounded CoT Skill Reasoner:} This module orchestrates the translation of perceptual inputs into executable actions by synthesizing bounding box coordinates with morphological features via the Qwen-vl reasoning engine. Instead of simple instruction following, we deploy a structured Chain-of-Thought (CoT) protocol to enforce physical grounding. The system is instantiated with a constraint-aware prompt: \enquote{\textit{You are a robot with five skills: Side Grasp, Lift Up, Top Pinch, Pour Water and Delivery. The image you are observing has a resolution of 640x480. Based on the observation, the coordinates of the bounding box and the shape information of the object, choose the skill without collision and generate an action plan based on these skills.}} By embedding task-specific primitives (e.g., Side Grasp, Pour Water) within this reasoning framework, the model explicitly maps latent geometric affordances to appropriate manipulation strategies. Crucially, this mechanism resolves operational ambiguities by integrating safety-critical constraints, prioritizing kinematically feasible and collision-free interactions based on environmental context. The final decision is derived from a confidence-weighted assessment of candidate skills, ensuring robust planning under uncertainty.

\noindent
\textbf{Skill Library}
The skill library encompasses a comprehensive set of predefined grasping configurations parameterized by finger joint angles, supporting three core manipulation primitives. Side grasping is employed when objects can be accessed laterally without obstruction. Lifting is utilized for vertical retrieval, often in contexts where obstacles limit horizontal approach. Pinching is reserved for handling thin or delicate items such as paper sheets and cables. These skills define hand posture without imposing object-specific constraints. The \gss{} selects the appropriate skill by combining geometric priors with real-time RGB observations, determining feasible hand configurations for the current scenario. Each skill corresponds to distinct model parameters, and the \rgmps{} subsequently generates corresponding action trajectories dynamically during real-time inference.

\subsection{Recursive Adaptive Spiking Network}
\label{subsec:wqkv}
\noindent{} 
\textbf{Motivation of \wqkv{}.} Effective robotic operation necessitates a precise understanding of spatial relationships within visual scenes, particularly the correspondence between image regions and the robotic end-effector position. Existing methods often face challenges in inferring such relational structures in novel environments, primarily due to limitations in modeling spatial relationships and dense representation learning that hinder generalization. To address this issue, we introduce the \wqkv{} framework. This framework adaptively learns spatial dependencies between the robot and target objects in previously unseen settings through recursive spiking feature processing, while enhancing robustness and mitigating overfitting by employing Gaussian action space modeling when training data is scarce.

Our \wqkv{} advances action generation through three interconnected technical innovations. An Adaptive Decay Mechanism (ADM) dynamically regulates the retention of historical context, enabling the model to capture long-range dependencies while maintaining spatial accuracy. Adaptive spiking neurons selectively amplify task-relevant visual features, such as object contours, and suppress non-essential background information within dense feature maps. These elements are unified by a hierarchical fusion framework, which seamlessly integrates multi-scale representations to preserve both high-level semantics and detailed spatial structure through learned attention weighting.

Specifically, the \wqkv{} comprises three principal components: (1) a Stem layer for preliminary feature extraction, (2) three hierarchical processing stages, each containing two \wqkv{} blocks, and (3) a multi-scale feature fusion module integrated with a GMM, as illustrated in Fig.~\ref{fig:pipeline}. Given an input RGB image \( \mathcal{O} \in \mathbb{R}^{H \times W \times 3} \), the Stem layer performs initial spatial downsampling and feature embedding, reducing the resolution to \(\frac{H}{4} \times \frac{W}{4}\) while preserving critical low-frequency information. This design significantly mitigates computational overhead for subsequent operations. The transformation can be formally expressed as: 
\begin{equation}\small
\label{eq:stem}
F_0 = \mathcal{P}(SReLU(BN(\mathcal{C}_{3\times3}(\mathcal{O})))),  
\end{equation}
where \(\mathcal{C}_{3 \times3 }(\cdot)\) denotes a \(3 \times3\) convolutional layer, \(BN(\cdot)\) refers to batch normalization, and \(SReLU(\cdot)\) represents the Squared ReLU activation function. This activation improves gradient flow and increases the network's nonlinear representational capacity. The operator \(\mathcal{P}(\cdot)\) corresponds to a max pooling layer. The resulting feature map \(F_0\) is then split into two parallel processing pipelines within the \wqkv{} block at Stage 1: one for dynamic recursive computation and another for spiking-based dense feature extraction.

\noindent 
\textbf{Dynamic Recursive Computation.} In this branch, \(F_0\) is first processed by the ADM to generate content-adaptive decay factors that modulate historical memory retention. This process extracts local spatial information through convolutional layers, applies channel-wise recalibration, and constrains the decay factor to the range \([0,1]\) via a sigmoid activation function, formulated as:
\begin{equation}\small
\label{eq:adm}
\mathcal{W} = \sigma(\mathcal{C}_{1\times1}(SReLU(\mathcal{C}_{3\times3}(F_0)))),  
\end{equation}
where \(\mathcal{C}_{1\times1}(\cdot)\) denotes a \(1\times1\) convolutional layer that performs channel-wise recalibration, and \(\sigma(\cdot)\) represents the sigmoid activation function. We then apply Rotary Position Embeddings (RoPE) to encode positional information via geometric rotations in feature space. This method enhances the capacity of model to capture position-sensitive dependencies. Unlike standard additive or learnable position embeddings, RoPE incorporates relative position awareness by applying rotational transformations to queries and keys during attention computation. This design offers two main advantages: it removes learnable position parameters, thereby reducing overfitting risk while preserving efficiency; and it encodes positional offsets as angular shifts, ensuring attention scores inherently reflect relative token distances. Formally, we first generate the keys \(K_s\) and values \(V_s\) from \(F_0\) via a \(1 \times 1\) convolution. To incorporate positional awareness, we apply the rotation matrix \(\mathbf{R}_{\Theta}\) to pairs of channels in \(K_s\). The rotation matrix is defined as:
\begin{equation}\small
\label{eq:rope}
\mathbf{R}_{\Theta}(x) = 
\begin{bmatrix}
\cos(\theta_{h,w,j}) & -\sin(\theta_{h,w,j}) \\
\sin(\theta_{h,w,j}) & \cos(\theta_{h,w,j})
\end{bmatrix},
\end{equation}
where \(\theta_{h,w,j}\) is the two-dimensional position encoding:
\begin{equation}\small
\theta_{h,w,j} = \theta_j \cdot x_h + \theta_j\cdot x_w ,\quad
\theta_j= \frac{1}{10000^{\frac{2j}{C}}} ,
\end{equation}
where \(x_h\) and \(x_w\) represent the discrete horizontal and vertical coordinates of a pixel, and the frequency parameter \(\theta\) controls the rotation speed across channel dimensions (indexed by \(j \in [0, C/2)\)). Using a base constant of 10000, low-frequency components emphasize global structure, while high-frequency components capture local positional details. By applying \(\mathbf{R}_{\Theta}(x)\) to the query and key representations, we ensure that their inner product encodes relative positional information. After positional encoding, \(K_s\) and \(V_s\) are partitioned into \(16 \times 16\) image patches. Sequential dependencies among these patches are modeled using a recurrent \(WKV\) mechanism:
\begin{equation}\small
\label{eq:wkv}
\begin{gathered}
        WKV_{i} = \frac{n_{i} + \exp(u) \odot k_{i} \odot v_{i}}{d_{i} + \exp(u) \odot k_{i}}, \\
        \underbrace{\tcbhighmath[colframe=white, boxsep=1pt, top=2pt, bottom=2pt]{n_{i} = n_{i-1} \odot \exp(-\mathcal{W}) + k_{i} \odot v_{i}}}_{\textbf{Cumulative memory of } k_{i} \odot v_{i}}, 
        \underbrace{\tcbhighmath[colback=green!8, colframe=white, boxsep=1pt, top=2pt, bottom=2pt]{d_{i} = d_{i-1} \odot \exp(-\mathcal{W}) + k_{i}}}_{\textbf{Cumulative memory of } k_{i}},
    \end{gathered}
\end{equation}
where \(i \in [0, (H \times W)/(16 \times 16))\) indexes the sequence of image patches, and \(\exp(\cdot)\) denotes the element-wise exponential function. \(k_i\) and \(v_i\) denote the corresponding patch representations extracted from \(K_s\) and \(V_s\). To ensure consistent accumulation, the initial states (at step \(i=0\)) are set as \(n_0 = k_0 \odot v_0\) and \(d_0 = k_0\). The term \(u \in (0,1)\) represents a learnable position bias, and \(\mathcal{W} \in \mathbb{R}^C\) is the channel-wise time-decay vector.

\noindent 
\textbf{Motivation of Spatial Memory.} The recurrent connections in our framework build a spatial memory of observed scenes for the robot. This memory supports locating end-effector positions that are most relevant to the current task. However, recurrent computation is prone to vanishing gradients, which complicates training and often requires large amounts of data to alleviate this issue. To overcome this issue, we introduce an Adaptive Decay Mechanism that dynamically adjusts the decay rate of historical memory, preserving essential spatial information while adaptively emphasizing task-critical image patches. Here \(\mathcal{W}\) denotes the content-adaptive decay factors that govern memory retention (\cref{eq:adm}). Finally, a gating factor \(R_s\) produces a dynamic weight to modulate how the output of the Spatial Mixing Block influences the current state:
\begin{equation}\small
F_{S0} = \sigma(\mathcal{C}_{1\times1}(F_0))\odot WKV
\end{equation}

Then we apply the Channel Mixing Block to $F_{S0}$ to reallocate channel-wise feature weights for feature extraction:
\begin{equation}\small
\label{eq:channel}
F_{C0} = \sigma(\mathcal{C}_{1\times1}(F_{S0}))\odot (SReLU(\mathcal{C}_{3\times3}(F_{S0})))
\end{equation}
\noindent 
\textbf{Spiking Dense Feature Extraction.} In the parallel branch, dense visual features are extracted using a Dense Block. These features capture contextual information and textural details of target objects. The input \(F_0 \in \mathbb{R}^{\frac{H}{4} \times \frac{W}{4} \times C}\) is processed by the Dense Block, yielding an enriched representation \(F_k \in \mathbb{R}^{\frac{H}{4} \times \frac{W}{4} \times 4C}\). This output then passes through a Linear layer with \(C/r\) output channels, producing two new feature maps \(B_k\) and \(D_k\). The map \(B_k\) is subsequently used to generate a spatial attention mask \( X_k \):
\begin{equation}
\label{spacial}
X_k = B_k \bar{I}_{k} B_{k}^{T},
\end{equation}
where \( \bar{I} = \frac{1}{n} \left( I - \frac{1}{n} \mathbf{1} \right) \) and \( n = h \times w \). \( I\) and \( \mathbf{1} \) are the \( n \times n \) identity matrix and matrix of all ones, respectively, and \( X_k \in \mathbb{R}^{n \times n} \). We define \( x_{(i,j)} \in X_k \) (where \( 0 \leq i < n \); \( 0 \leq j < n \)), which encodes the global dependencies between the \( i \)-th pixel and the \( j \)-th pixel from \( B_k \). \( X_k \) is fed into a softmax function and then multiplied by \( D_k \):
\begin{equation}
    U_{k}=\operatorname{softmax}\left(X_{k}\right) D_{k},
\end{equation}

Then, $U_k $ is reshaped and transposed to $ \mathbb{R}^{n \times C/r} $. Subsequently, it is fed to a $ 1 \times 1 $ convolutional layer and added to the spiking feature $ \mathcal{A}(F_k) $ by our adaptive spike neuron:
\begin{equation}
\label{psnl}
    Q_A=\mathcal{C}_{1\times1}\left(U_{k}\right)+\mathcal{A}\left(F_{k}\right).
\end{equation}

\noindent 
\textbf{Motivation of Adaptive Spike Neuron.} Current robotic manipulation models typically generate actions based solely on spatial perception, overlooking temporal information. To incorporate temporal reasoning into spatial feature understanding, we employ a spiking neuron mechanism. Conventional spiking neurons combine spatial information from previous timesteps with a decay factor to modulate the current signal, enabling robots to integrate observations from past frames when processing current visual input. However, traditional spiking neurons rely on a fixed decay factor and a binary signaling mechanism, which introduce training difficulties in spiking neural networks due to non-differentiable operations during gradient backpropagation. To address these issues, we introduce an adaptive spiking neuron. Our design replaces the static decay factor with a dynamically adjustable one at each timestep. Training focuses on learning this adaptive decay factor rather than directly optimizing the spiking neural network parameters. This approach mitigates the refractory period characteristic of spiking neurons and resolves gradient propagation challenges posed by non-differentiable components, resulting in a more stable and efficient learning process. The governing equations of our adaptive spike neuron are detailed as: 
\begin{equation}
\begin{gathered}\label{eq:spike}
V_t = H_{t-1} + X_t, \quad S_t = \Theta(V_t - u_{th}), \\
H_t = V_t e^{-\Delta t/\tau} (1 - S_t) + V_{reset} S_t,
\end{gathered}
\end{equation}
where \(\Delta t\) denotes the time interval (typically unit time), and \(V_t\) represents the membrane potential obtained through integration of spatial input \(X_t\) and temporal input \(H_{t-1}\). Here $X_t$ is extracted from dense visual features, and $H_{t-1}$ corresponds to the membrane potential from the previous timestep after undergoing reset or spike emission. If the membrane potential exceeds the threshold $u_{th}$, the neuron fires a spike; otherwise, it remains silent. The function $\Theta(\cdot)$ denotes the Heaviside step function, defined as $\Theta(x)=1$ for $x \geq 0$ and $\Theta(x)=0$ otherwise, yielding a binary output $S_t \in \{0,1\}$. The term $H_t$ indicates the temporal output, where $\tau$ represents the adaptive decay factor produced by our channel-wise attention mechanism. The reset potential $V_{\text{reset}}$ is engaged after a spike is fired. In the absence of firing, $V_t$ decays exponentially according to the adaptive factor $\tau$. To create an adaptive decay factor $\tau$ suited for particular tasks, the computation of $\tau$ is seamlessly incorporated into the gradient backpropagation process of the policy:
\begin{equation}
    \tau=\mathbb{E}[\mathcal{C}_{1\times1}\left(R(\mathcal{C}_{1\times1}(F_{k}))\right)],
\end{equation}
where $\mathbb{E}[\cdot]$ computes the average of mapped values across channels, and $R(\cdot)$ denotes a feature reshaping operation. This integration enables the decay factor to adjust dynamically based on the complexity of the current motion sequence. Finally, we apply a \(1\times1\) convolutional layer to \(Q_{A}\) to obtain query vectors \(Q\) with matching channel dimensions to the key and value vectors \(K\) and \(V\). A guided self-attention mechanism then merges the dynamically recurrent features and the spiking-based dense features as follows:
\begin{equation}\small
\label{eq:GSNET}
G = \mathrm{softmax}\left( \frac{Q K^{\top}}{\sqrt{d_k}} + B \right) V,
\end{equation}
where \(d_k\) denotes the dimensionality of the queries and keys, and \(B\) represents a relative position bias. The feature map \(F_1\) is produced by down-sampling the output from two successive \wqkv{} blocks using a \(3 \times 3\) convolution. This procedure is repeated in Stage 2 and Stage 3. The resulting multi-scale features \(F_1\), \(F_2\), and \(F_3\) are subsequently fused through a set of learnable weights:
\begin{equation}\small
F_f =\alpha_1(\mathcal{C}_{1\times1}(F_{1}))+\alpha_2(\mathcal{C}_{1\times1}(Up(F_{2})))+\alpha_3(Up(F_{3})),
\end{equation}
where \(F_i\) represents the feature map produced at Stage \(i\) (with \(i = 1, 2, 3\)), and \(\alpha_i\) denotes learnable parameters that weight each level during feature fusion. The fused feature map \(F_f\) is then used to generate an initial action prediction:
\begin{equation} \small
\label{eq:ain}
    a_{in} = \mathrm{Linear}\bigl(\mathcal{C}_{3\times3}(F_f)\bigr),
\end{equation}

To train the model, we minimize the mean-squared error between the predicted action and the ground-truth action from human demonstrations using the following loss:
\begin{equation}
\label{eq:loss}
    \mathcal{L} = \mathrm{MSE}\bigl( a_{in},\; a_{\mathrm{ground}} \bigr),
\end{equation}
where \(\mathcal{L}\) is the loss and \(a_{\mathrm{ground}}\) is the ground-truth action.

\begin{algorithm}[!t]
\caption{The \rgmps{} Framework}
\label{alg:RGMP}
\SetNoFillComment
\SetArgSty{textnormal}
\small{\KwIn{Training epochs $E$, conversation round $T$, human instructions $\mathcal{I}$, human demonstrations $\mathcal{D}$ with capacity $M$, current observation $\mathcal{O}$, VLM model $\mathcal{W}$, Stage of \rgmps{} $\mathcal{S}$, \rgmps{} $\mathcal{G}_{m}$ }}
\small{\KwOut{Robotic actions $a^{*}$}}

 \BlankLine
{\tcc{Human Demonstration Collecting:}}
\For {$i=1, 2, ..., M$ }{
    $d_{i}\leftarrow(\mathcal{O}_{i}, \mathcal{J}_{i})$ through {\cref{eq:collection}}  {{\tcp*{\cref{subsec:data}}}}
    
    $\mathcal{D} \leftarrow \mathcal{D} \cup \{d_{i}\}$  

    $\Theta \leftarrow \mathrm{EM}(\mathcal{D})$ via \cref{eq:GMM} 
    }  
    return $\mathcal{D}$, $\Theta$
    
\BlankLine


{\tcc{\rgmps{} Training pipeline:}}
\For {$e=1, 2, ..., E$ }{
    \BlankLine
    
    $F_{0}\leftarrow Stem(\mathcal{O}_{i})$ by \cref{eq:stem} {{\tcp*{\cref{subsec:wqkv}}}}

    ${\mathcal{W}},{K_s}, {V_s}  \leftarrow \mathcal{A_d}(F_{0})$, $\mathcal{R}(F_{0})$ by \cref{eq:adm}, \cref{eq:rope}

    ${F_{1},F_{2},F_{3}}  \leftarrow \mathcal{S}({K_s}, {V_s}, {\mathcal{W}})_{\times 3}$

    ${a_{in}}  \leftarrow \mathcal{M}(F_{1},F_{2},F_{3})$ by \cref{eq:ain}

    $\mathcal{L} \leftarrow  ({a_{in}},{a_{ground}})$ through \cref{eq:loss} 

    $\mathcal{G}_{m}  \leftarrow \mathcal{G}_{m}  - \eta \nabla \mathcal{L}$
    }
    return $\mathcal{G}_{m}$
 \BlankLine

{{\tcc{Inference pipeline:}}}

\For {$t=1, 2, ..., T$ }{

    \BlankLine
    $ Box(x_1,y_1,x_2,y_2) \leftarrow \mathcal{Q}(\mathcal{I}, \mathcal{O}| \mathcal{C})$ by Eq.~({\ref{eq:vlm}})

    $\mathcal{O}_{s} \leftarrow  (x_1,y_1,x_2,y_2),\kern 2pt \phi_{a}(\mathcal{O},Box)$

    
   $\mathcal{S}_{seq}, \mathcal{P} \leftarrow \mathcal{Q}_{CoT}(\mathcal{I}, \mathcal{O}_{s}| \mathcal{C})$  
   
    Voice $\leftarrow$ response in $\mathcal{P}$
    
    \BlankLine
    \If {$\mathcal{S}_{seq} \neq \emptyset$}{
    \For{ $s_k$ in $\mathcal{S}_{seq}$}{



            $a^{*}\leftarrow \mathcal{G}_{m}(\mathcal{O}, s_k, \Theta)$ through \cref{eq:regression}



        }

        Execute $a^{*}$}
}
\end{algorithm}

\begin{figure*}[!ht]
\centering
  \includegraphics[width=0.8\textwidth]{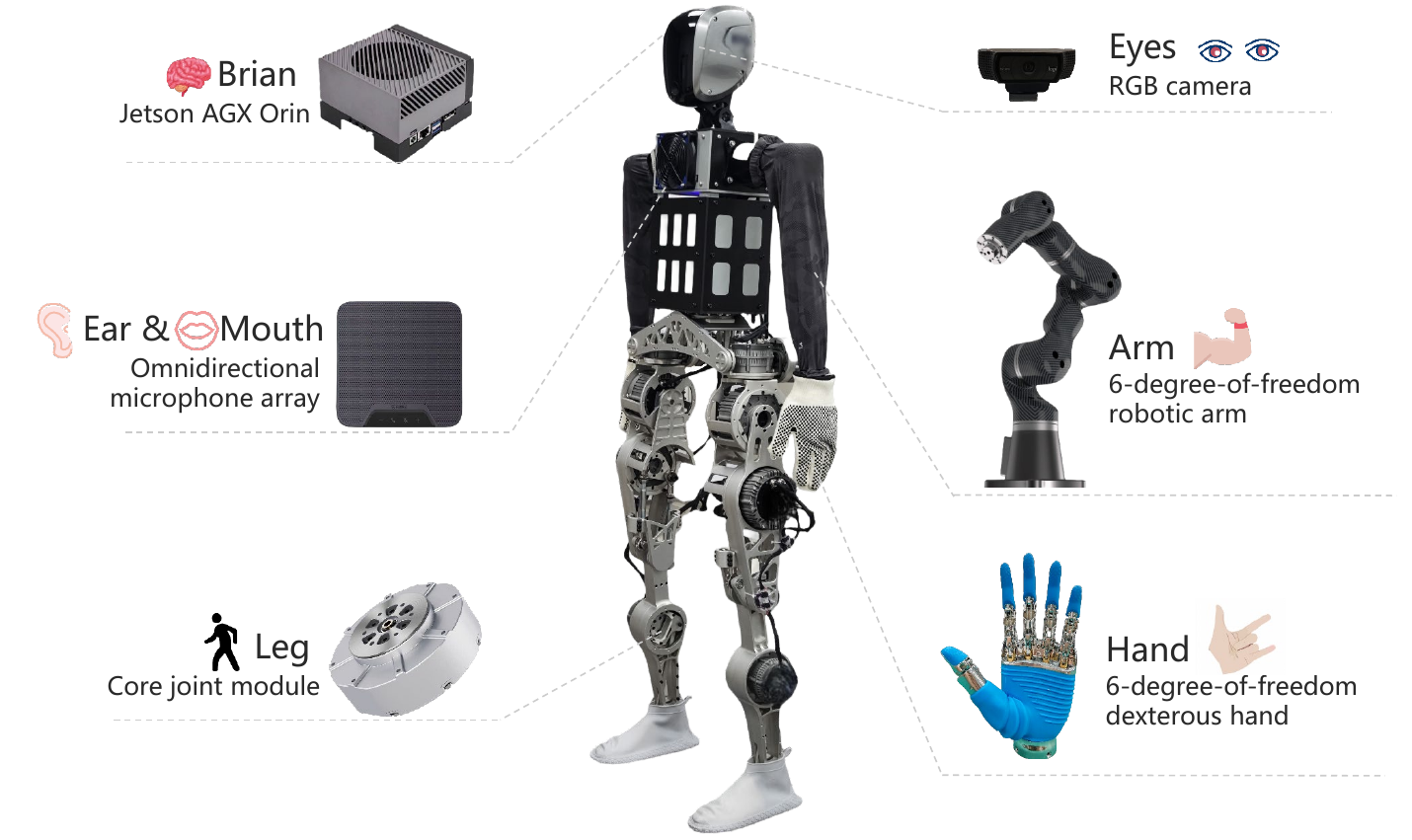}
  \vspace{-0.4cm}
\caption{\textbf{Hardware configuration of our humanoid robot.} The platform integrates several core subsystems: an NVIDIA Jetson AGX Orin module for onboard computation; a head-mounted RGB camera for egocentric vision; an omnidirectional microphone array for audio input; a 6-degree-of-freedom robotic arm paired with a dexterous hand for manipulation; and a multi-axis joint module for torso mobility. These components collectively enable environmental perception, physical interaction, and autonomous movement. See \cref{sec:ExperSetup} for details.}
\label{fig:hard}
\vspace{-0.2cm}
\end{figure*}

\noindent 
\textbf{Motivation of Gaussian Mixture Model.}
A single Gaussian distribution~\cite{chi2023diffusion} often regresses to the mean, which can suppress distinct action modes and compromise control precision. To better represent the multi-modal nature of robotic actions, we adopt a Gaussian Mixture Model (GMM). Let \(\mathbf{x} \in \mathbb{R}^n\) denote the ground-truth joint configuration. The GMM with \(K=6\) components is defined as:
\begin{equation}\small
\label{eq:GMM}
P(\mathbf{x} \mid \Theta) = \sum_{k=1}^{K} \alpha_{k} \, \mathcal{N}\left(\mathbf{x} \mid \mu_{k}, \Sigma_{k}\right),
\end{equation}
where \(\alpha_k\), \(\mu_k\), and \(\Sigma_k\) denote the prior weight, mean, and covariance of the \(k\)-th component. These parameters are estimated via the expectation-maximization algorithm to maximize the data likelihood. To refine the initial prediction \(a_{\text{in}}\), we compute the Mahalanobis distance to each GMM component. As the intrinsic metric for Gaussian distributions, it utilizes covariance to normalize deviations, ensuring statistical consistency by penalizing errors in low-variance dimensions:
\begin{equation}\small
l_{k} = \sqrt{\left(a_{\text{in}} - \mu_{k}^{\omega}\right)^{\top} \left(\Sigma_{k}^{\omega, \omega}\right)^{-1} \left(a_{\text{in}} - \mu_{k}^{\omega}\right)}.
\end{equation}

The final action \(a^{*}\) corresponds to the mean vector of the component with the smallest distance:
\begin{equation}\small
\label{eq:regression}
a^{*} = \arg \min_{\mu_{k}^{\omega}} \, l_{k}.
\end{equation}

The pseudocode of \rgmps{} is presented in Algorithm~\ref{alg:RGMP}.

\begin{figure*}[t]
  \includegraphics[width=\textwidth]{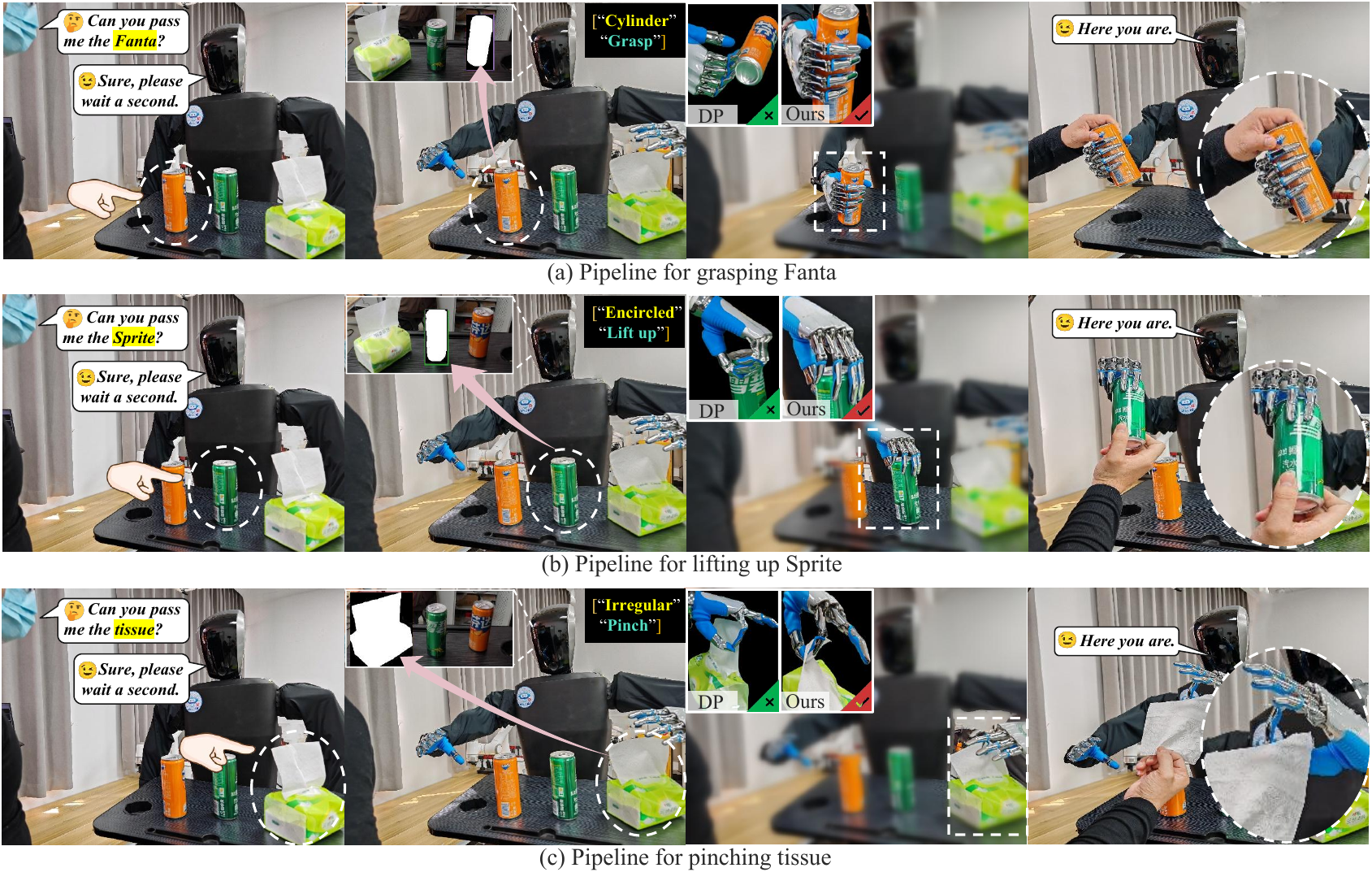}
  \vspace{-0.9cm}
\caption{\textbf{Experimental validation within the interactive bar service scenario.} Evaluation of the \rgmps{} framework regarding the delivery of beverages and tissues. The training phase involves 40 expert trajectories per object category. Quantitative results demonstrate that \rgmps{} attains superior success rates and robustness relative to the Diffusion Policy (DP) baseline. See \cref{sec:sota} for details.}
\label{fig:experiment1}
\vspace{-0.4cm}
\end{figure*}

\section{Experiments}
This section presents a systematic experimental evaluation of the proposed \rgmps{} framework. Our experiments aim to validate the effectiveness and generalization capacity of the overall architecture and its two core modules: the \gss{} and the \wqkv{}. We analyze the distinct contribution of each module and compare the full framework against several state-of-the-art baselines on a range of manipulation tasks. The following sections detail the experimental setup, implementation specifics, evaluation metrics, and present a comprehensive discussion of the results.
\subsection{Experimental Setup}
\label{sec:ExperSetup}
\noindent \textbf{Hardware Setup}.
\label{exp:setup}
Our experiments are conducted on three distinct robotic platforms. The primary platform is an upper-body humanoid robot with four integrated subsystems: vision, audio, decision-making, and actuation. A head-mounted egocentric RGB camera provides primary visual input. An iFLYTEK S0Y22F omnidirectional microphone and speaker array, mounted on the torso, handles voice interaction. An onboard NVIDIA Orin compute module performs high-level behavioral planning. The actuation system consists of two 6-degree-of-freedom (6-DOF) robotic arms, each equipped with a 6-DOF dexterous hand. The entire system (as shown in \cref{fig:hard}) is self-powered and communicates via a local wireless network, enabling fully untethered operation. To assess cross-embodiment generalization, we also employ a desktop dual-arm manipulation robot. This secondary platform is equipped with an overhead RGB camera and two 6-DOF robotic arms dedicated to executing precise physical manipulation tasks. The third platform comprises the Aloha robot, which incorporates a centrally positioned RGB camera and dual robotic arms equipped with wrist-mounted RGB sensors.

\noindent \textbf{Dataset and Evaluation Criteria}.
\label{subsec:data}
To verify the efficacy of the proposed \rgmps{} framework, we curated a comprehensive dataset comprising 500 expert trajectories within the skill library. Each trajectory constitutes an operational path paired with a visual observation captured immediately before the initiation of a robotic task. In particular, the visual input facilitates the mapping to a specific action, while the trajectory delineates the transition of the manipulator from the starting configuration to the target spatial coordinates and end-effector pose. This data collection process is formulated in the following manner.
\begin{equation}
\label{eq:collection}
d_{i}=(\mathcal{J,O}),
\end{equation}
where $\mathcal{J}$ signifies the joint space of the robotic system and $\mathcal{O}$ represents the corresponding perceptual observation. During physical experiments, the system performance is quantified via two complementary metrics. The skill selection accuracy $Acc_s$ represents the capability of the model to identify the appropriate primitive for a given instruction. Furthermore, the execution success rate $Acc_t$ evaluates the precision with which the manipulator interacts with the target object. Consequently, the comprehensive success rate $Acc$ is defined as the product of these two components.
\begin{equation}
\label{eq:acc}
Acc = Acc_s \times Acc_t
\end{equation}

\noindent \textbf{Real-world Task Configurations.}
To verify the performance of the \rgmps{}, we conduct a comprehensive comparative evaluation against several state-of-the-art baselines across 10 manipulation tasks. This evaluation suite comprises five simulation tasks from the ManiSkill2 benchmark and five physical experiments. The specific configurations of the real-world tasks are outlined in the following descriptions.

\textbf{\textit{Interactive Bar Service}.}
In this scenario, the humanoid robot engages in verbal dialogue with a human and provides assistance based on vocal instructions. These tasks include the delivery of beverages or tissues (as shown in \cref{fig:experiment1}). During evaluation, the skill selection accuracy $Acc_s$ is recorded as 0 if the robot selects an inappropriate skill. Typical failures primarily involve unstable side grasps on deformed objects and collisions with peripheral obstacles during cluttered manipulation. Conversely, successful skill identification is denoted as $Acc_s = 1$. Furthermore, the execution success rate $Acc_t$ is recorded as 1 only upon the successful grasping of the target object. For each behavioral primitive, the model is trained using 40 expert trajectories and subsequently validated across 20 randomly positioned objects, yielding the mean success rate as the final accuracy.

\begin{figure*}[t!]
  \includegraphics[width=\textwidth]{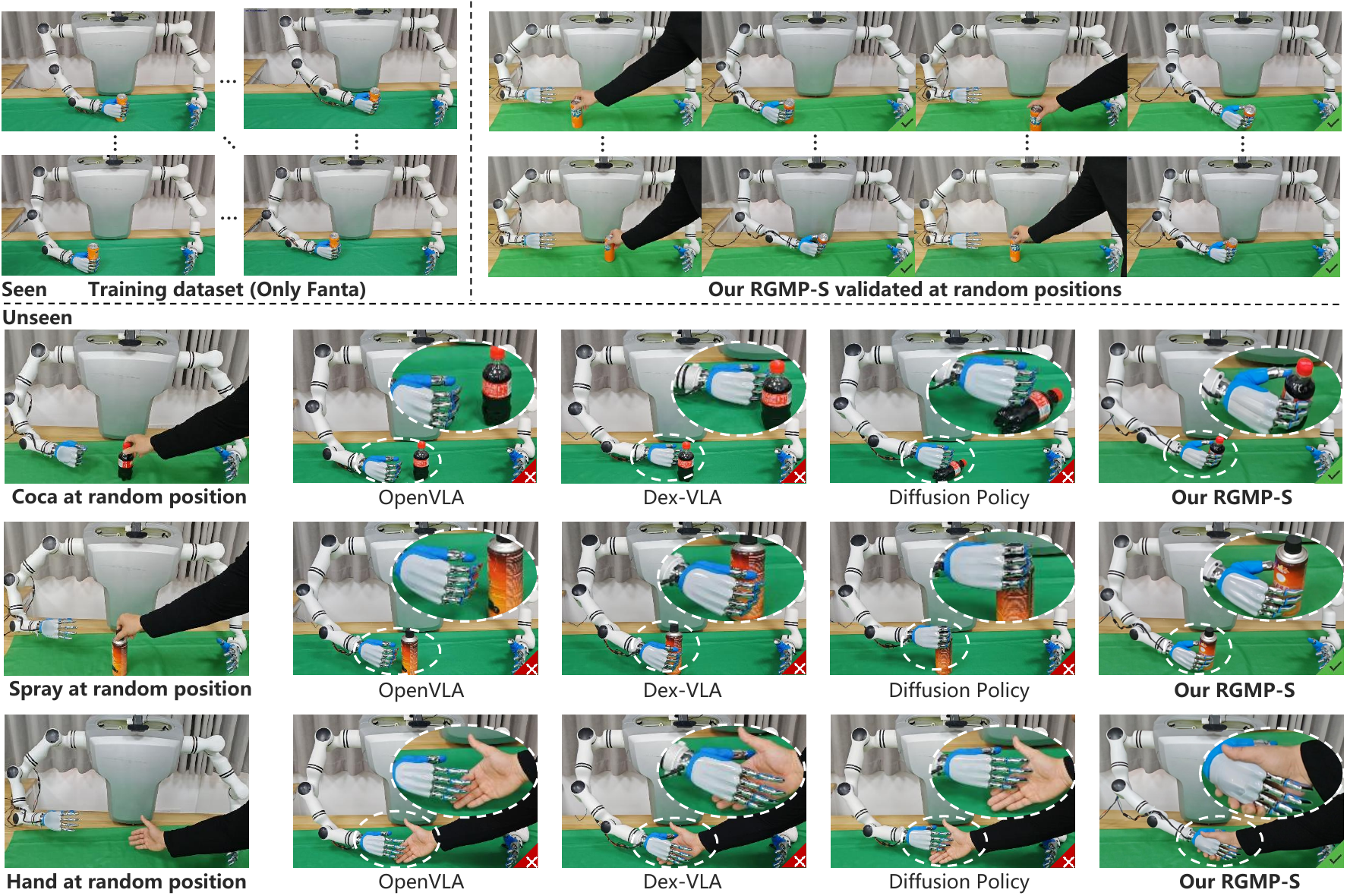}
  \vspace{-0.8cm}
  \caption{\noindent \textbf{Generalized manipulation capabilities of the \rgmps{}.} 
We evaluate the \rgmps{} regarding its ability to grasp various novel objects at randomized positions. The training phase relies on a dataset containing only 40 expert demonstrations focused on grasping a Fanta can. Experimental results indicate that \rgmps{} achieves proficient success rates when grasping the Fanta can across diverse spatial configurations. Furthermore, the system demonstrates substantial zero-shot transfer to previously unencountered entities such as a Coke can, a spray bottle, and a human hand. This outcome highlights the inherent versatility and adaptability of the architecture across diverse manipulation scenarios. See \cref{sec:sota}.}
\label{fig:experiment2}
\vspace{-0.4cm}
\end{figure*}

\textbf{\textit{Zero-shot Grasping}.}
To assess the generalization capability of the \rgmps{} framework within previously unseen environments (as shown in \cref{fig:experiment2}), we evaluate the performance of the action generation ability. The training phase utilizes a constrained dataset comprising 40 expert trajectories solely for grasping a Fanta can. During the inference stage, the zero-shot transfer capability is validated through grasping trials performed at randomized positions on novel objects with distinct geometries including a Coke can, a spray bottle, and a human hand. In addition to testing unseen categories, we examine spatial generalization by deploying the originally seen Fanta can in various randomized positions across the workspace. Each evaluation trial is quantified via a binary success metric, where a stable retrieval of the target is recorded as 1 and a failure as 0. For each object category, we conduct 100 independent trials at randomized locations to determine the final success rate.

\textbf{\textit{Towel Folding}}. The towel folding task involves the manipulation of deformable objects through a sequence of maneuvers intended to achieve a predefined configuration. This environment involves high degrees of freedom and stochastic physical properties including randomized initial poses and various fabric textures. Success is defined as aligning the fabric edges within a specified spatial tolerance ($\pm$ 2 cm) and maintaining a stable folded geometry. The training corpus for this task comprises 100 expert trajectories that encompass diverse towel dimensions and colors. For evaluation, we employ a validation scheme involving randomized placements and novel towel instances to assess the generalization performance of the proposed framework.

\textbf{\textit{Pouring Water}}. The primary objective of this task is the coordination of a source container to facilitate the successful transfer of liquid into a target receptacle. The environment involves complex fluid dynamics and stochastic physical properties including randomized initial positions and various container geometries. Success is defined as the spill-free transfer of water into the target receptacle. The training corpus for this task comprises 100 expert trajectories that encompass different initial fill levels and container types. For evaluation, we employ a validation scheme involving novel receptacles and randomized spatial configurations to assess the robust generalization capability.

\begin{figure*}[!ht]
  \includegraphics[width=\textwidth]{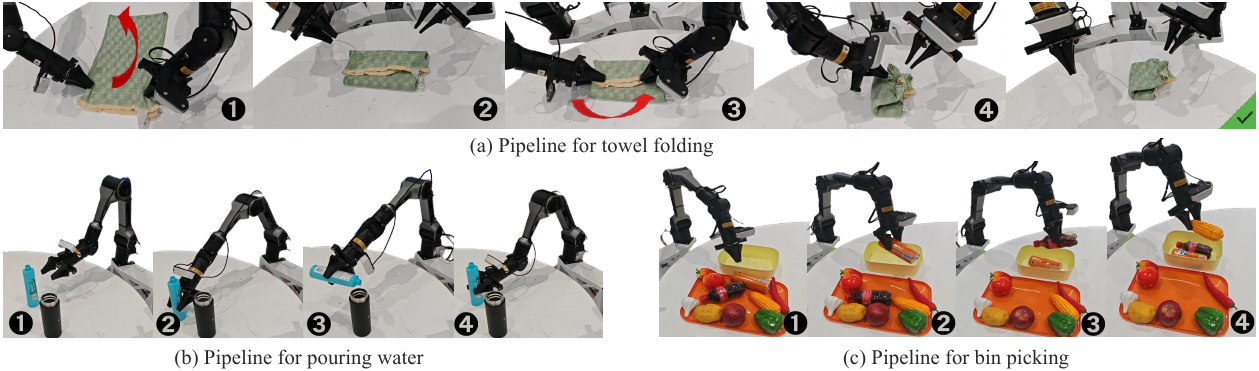}
  \vspace{-0.8cm}
\caption{\textbf{Validation on long-horizon tasks.} Evaluation of the \rgmps{} framework regarding towel folding, water pouring, and bin picking. See \cref{sec:sota}.}
\label{fig:experiment3}
\vspace{-0.2cm}
\end{figure*}

\textbf{\textit{Bin Picking}}. Within the bin picking scenario, the robot is tasked with retrieving objects from a storage bin and relocating them to a designated target zone under cluttered conditions. The environment involves complex occlusions and stochastic physical arrangements including randomized orientations and various object geometries within a confined space. Success is defined as the accurate retrieval of the target object and its stable placement within the goal area without unintended collisions. The training corpus for this task comprises 100 expert trajectories per object category to capture diverse grasping configurations. For evaluation, we employ a validation scheme involving novel object instances and randomized clutter patterns to assess the generalization robustness of the proposed framework (as shown in \cref{fig:experiment3}).

\begin{figure}[!t]
  \includegraphics[width=\columnwidth]{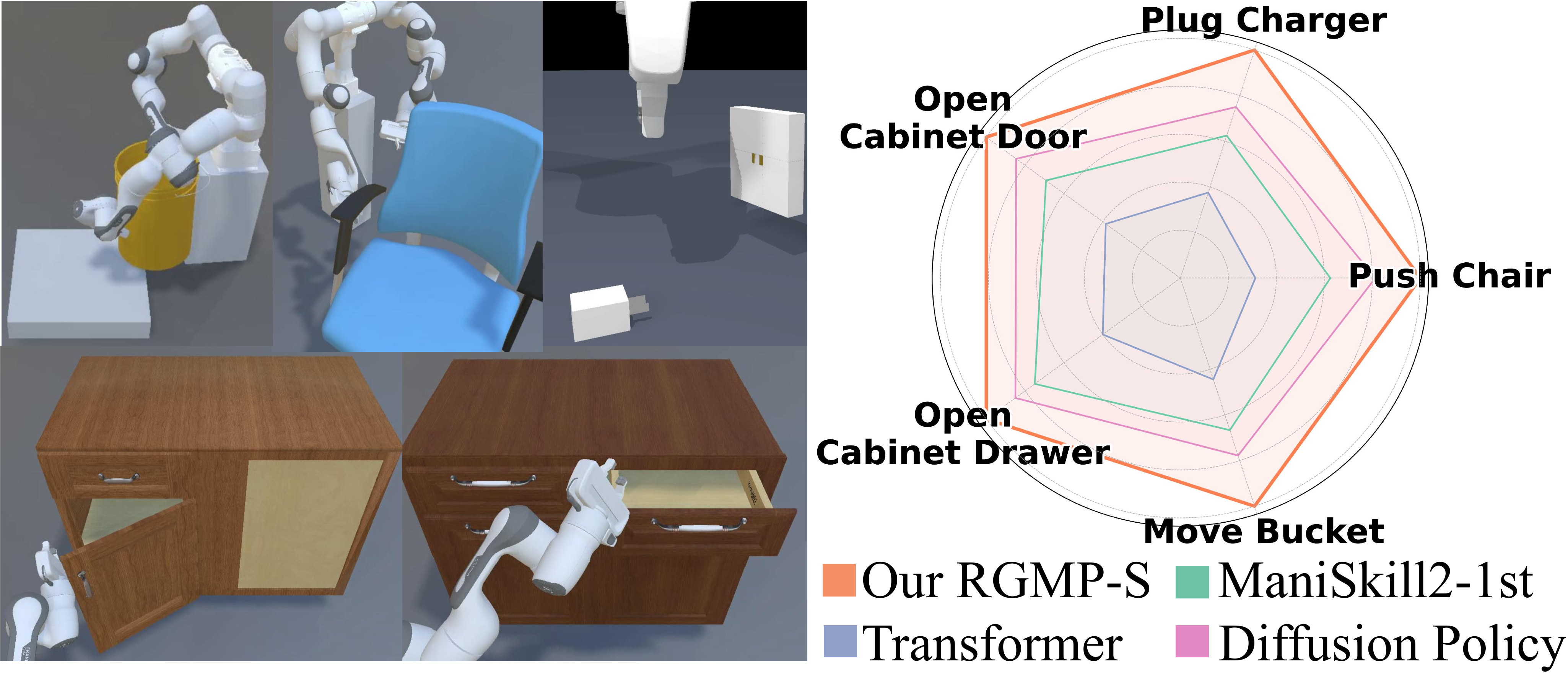}
  \vspace{-0.8cm}
  \caption{\textbf{Performance within the ManiSkill2 benchmark.} Quantitative evaluations on 5 representative manipulation tasks to compare the proposed framework with state-of-the-art baselines. Refer to \cref{sec:sota}.}
\label{fig:mani}
\vspace{-0.4cm}
\end{figure}

\noindent \textbf{Simulation Benchmark Specifications.}
\label{ap:mani}
To evaluate the proposed framework, we select five representative tasks from the publicly available and standardized ManiSkill2 simulation benchmark (as shown in \cref{fig:mani}). These tasks demand high-precision interaction and robust spatial reasoning. The detailed specifications for each task are described as follows.

 \textit{\textbf{OpenCabinetDrawer / CabinetDoor}} : These articulated object manipulation tasks require a single-arm mobile manipulator to actuate target components on a cabinet. These environments involve stochastic physical properties, including randomized friction and damping parameters for the joints. Success is defined as opening the target component to at least 90\% of the maximum range and maintaining a static state. The training corpus for these tasks comprises 300 expert trajectories per target, spanning 25 and 42 unique cabinets for drawers and doors, respectively. For evaluation, we employ a validation scheme involving ten unseen cabinets to assess the generalization capability of the model.

 \textit{\textbf{PushChair}} \& \textit{\textbf{MoveBucket}}: These tasks evaluate the capacity for precise bimanual coordination. In \textit{PushChair}, a dual-arm robot must translate a swivel chair to a target location without causing the object to tip over. Success requires the chair to remain static within 15 cm of the destination. The \textit{MoveBucket} task involves a complex manipulation sequence where the robot must lift a bucket containing a ball onto a raised platform without losing the internal object. Both tasks leverage a dataset of 300 trajectories per object, featuring 26 chairs and 29 buckets for training.

 \textit{\textbf{PlugCharger}}: This insertion task necessitates high-precision control of the end-effector. The manipulator must insert a charger into a wall receptacle. This task provides 1,000 successful trajectories for training. The validation scheme consists of two stages of 100 episodes, each featuring randomized initial joint configurations of the robot and varied initial poses of the charger relative to the wall.

\noindent \textbf{Counterparts.} {\color{black}{
We compare the proposed \rgmps{} framework against several representative baselines to evaluate performance across diverse manipulation tasks.

\begin{itemize}  
\item \textbf{ResNet} \pub{CVPR'16} {\cite{he2016deep}}: Robust visual feature extraction via deep residual learning for manipulation tasks.
\item \textbf{Transformer} \pub{NeurIPS'17} {\cite{vaswani2017attention}}: Global dependency modeling in trajectory sequences through self-attention.
\item \textbf{Diffusion Policy} \pub{RSS'23} {\cite{chi2023diffusion}}: Robotic action distribution representation via an iterative denoising process.
\item \textbf{ManiSkill2 1st} \pub{RSS'23} {\cite{gao2023two}}: A finetuning mechanism that achieved 1st place in the ManiSkill2 challenge.
\item \textbf{Octo} \pub{RSS'24} {\cite{team2024octo}}: An open-source generalist policy trained on massive datasets for robotic control.
\item \textbf{OpenVLA} \pub{CoRL'24} {\cite{kim2024openvla}}: Integration of VLMs with action tokens for open-world manipulation.
\item \textbf{RDT-1B} \pub{arXiv'24} {\cite{liu2024rdt}}: Performance scaling in multiple robotic tasks via a diffusion transformer architecture.
\item \textbf{Dex-VLA} \pub{CoRL'25} {\cite{wen2025dexvla}}: Alignment of vision and language representations for dexterous manipulation.
\end{itemize}

\subsection{Performance Comparison with SOTA Methods}
\label{sec:sota}
Comprehensive evaluations are conducted against state of the art baselines across ten manipulation tasks, encompassing five physical scenarios and five simulation benchmarks.
\noindent \textbf{Human-robot Interaction Analysis}. 
Table \ref{tab:bar} details the quantitative outcomes of the interactive bar service task. The tabulated data reveals that the absence of geometric guidance compromises the skill selection accuracy $Acc_s$ of the standard Qwen-vl baseline, particularly when paired with the ResNet50 backbone. This limitation highlights the necessity of geometric inductive biases for precise decision-making. In contrast, the integration of our \gss{} framework consistently outperforms the Qwen-vl counterpart across all diverse backbones. For example, within the Diffusion Policy configuration, \gss{} outperforms the baseline by 15\% on the squashed Coke task, which is calculated through the enhancement of overall accuracy from 0.40 to 0.55. This improvement primarily stems from the accurate identification of object orientations, which effectively mitigates unstable side grasps on deformed objects.

\begin{figure}[t]
  \includegraphics[width=\columnwidth]{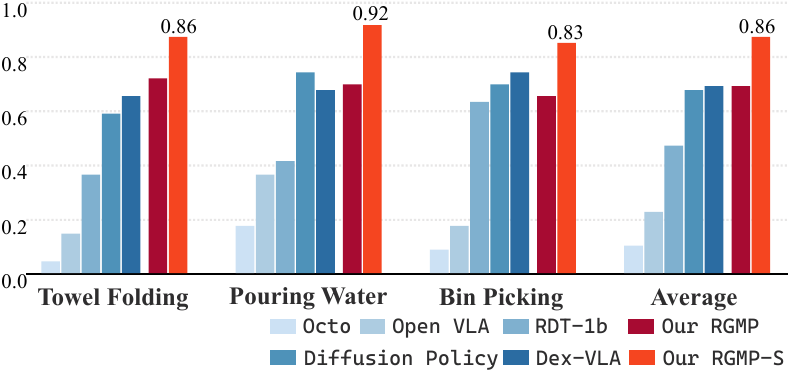}
  \vspace{-0.8cm}
\caption{\textbf{Quantitative analysis of long-horizon tasks.} The proposed framework is compared against state of the art methods for towel folding, water pouring, and bin picking. See \cref{sec:sota} for details.}
\label{fig:long}
\vspace{-0.4cm}
\end{figure}

\noindent \textbf{Generalized Performance Analysis}. 
We conduct extensive comparative experiments in Table \ref{table: grasp} with representative state-of-the-art models on the zero-shot grasping scenario, and our \rgmps{} achieves consistently superior performance. As depicted in the comparison, standard data-driven methods such as Octo and OpenVLA struggle to effectively generalize to novel categories like spray bottles and human hands due to the lack of explicit geometric reasoning. \rgmps{} consistently outperforms all other counterparts on different targets. Specifically, \rgmps{} outperforms the best baseline Dex-VLA by 12\% in terms of average success rate. Notably, in the challenging human hand handover task, our method attains a success rate of 0.93 compared to 0.84 for the second-best method, demonstrating the remarkably robust zero-shot transfer capability of the framework even when trained on a highly constrained dataset. Moreover, regarding computational efficiency, \rgmps{} achieves a high inference frequency of 75.2 Hz, significantly surpassing the computationally intensive models (e.g., 3.6 Hz for OpenVLA and 1.01 Hz for Diffusion Policy), thereby ensuring the low latency required for real-time robot control.

\begin{table*}[t]\small
\centering
\caption{\textbf{Ablation study regarding the \gss{} framework and Qwen-vl.} Experimental validations are conducted within environments featuring Fanta cans, Sprite cans, tissue paper, squashed Coke cans, and human hands with randomized object configurations for each trial. The training phase for each skill category involves 40 expert trajectories, while quantitative results are derived from 20 randomized validation trials. Comparative experiments are performed against representative baselines, including ResNet \cite{he2016deep}, Transformer \cite{vaswani2017attention}, ManiSkill2-1st \cite{gao2023two}, and Diffusion Policy \cite{chi2023diffusion}. The superior performance within each category is emphasized in \textbf{bold} and the second-best result is denoted with an \underline{underline}. See \cref{sec:ablation}.}
\label{tab:bar}
\vspace{-10pt}
{
    \resizebox{0.95\linewidth}{!}{
    \begin{tabular}{r||cccIcccIcccIcccIccc}
\hline\thickhline
\rowcolor{mygray}
        & \multicolumn{3}{cI}{Fanta} & \multicolumn{3}{cI}{Sprite} & \multicolumn{3}{cI}{Tissue} & \multicolumn{3}{cI}{Squashed Coke} & \multicolumn{3}{c}{Human Hand} \\
        \cline{2-16}
        \rowcolor{mygray}
      \multirow{-2}{*}{Methods}  & $Acc_s$ & $Acc_t$ & $Acc \uparrow$ & $Acc_s$ & $Acc_t$ & $Acc \uparrow$ & $Acc_s$ & $Acc_t$ & $Acc \uparrow$ & $Acc_s$ & $Acc_t$ & $Acc \uparrow$ & $Acc_s$ & $Acc_t$ & $Acc \uparrow$ \\
\hline\hline
\multicolumn{16}{l}{ with \textit{ResNet50}} \\
\hline
Qwen-vl \cite{bai2023qwen} & 0.65 & 0.54 & 0.35 & 0.60 & 0.42 & 0.25 & 0.65 & 0.46 & 0.30 & 0.65 & 0.46 & 0.30 & 0.70 & 0.57 & 0.40 \\
Our \gss{} & 0.85 & 0.53 & 0.45 & \underline{0.75} & 0.46 & 0.35 & 0.85 & 0.47 & 0.40 & 0.85 & 0.47 & 0.40 & 0.85 & 0.56 & 0.48 \\
\hline\hline
\multicolumn{16}{l}{with \textit{Transformer}} \\
\hline
Qwen-vl \cite{bai2023qwen}& 0.60 & \underline{0.58} & 0.35 & 0.65 & 0.54 & 0.30 & \underline{0.70} & 0.50 & 0.35 & 0.60 & 0.58 & 0.35 & 0.65 & 0.62 & 0.40 \\
Our \gss{} & 0.80 & 0.56 & 0.45 & \underline{0.75} & 0.53 & 0.40 & 0.85 & 0.53 & 0.45 & 0.85 & 0.53 & 0.45 & 0.85 & 0.64 & 0.54 \\
\hline\hline
\multicolumn{16}{l}{with \textit{ManiSkill2-1st}} \\
\hline
Qwen-vl \cite{bai2023qwen}& \underline{0.70} & 0.57 & 0.40 & 0.65 & 0.69 & 0.45 & 0.65 & 0.53 & 0.34 & 0.65 & 0.54 & 0.35 & 0.65 & 0.62 & 0.40 \\  
Our \gss{} & 0.85 & 0.53 & 0.45 & 0.80 & 0.68 & \underline{0.54} & 0.85 & 0.53 & \underline{0.45} & \underline{0.80} & 0.56 & \underline{0.45} & \underline{0.85} & 0.70 & \underline{0.60} \\
\hline\hline
\multicolumn{16}{l}{with \textit{Diffusion Policy}} \\
\hline
Qwen-vl \cite{bai2023qwen}& 0.65 & 0.76 & \underline{0.49} & 0.65 & \underline{0.75} & 0.50 & 0.65 & \underline{0.68} & 0.44 & 0.65 & \underline{0.62} & 0.40 & 0.70 & \underline{0.71} & 0.50 \\ 
Our \gss{} & \textbf{0.85} & \textbf{0.76} & \textbf{0.65} & \textbf{0.80} & \textbf{0.77} & \textbf{0.62} & \textbf{0.85} & \textbf{0.69} & \textbf{0.59} & \textbf{0.85} & \textbf{0.65} & \textbf{0.55} & \textbf{0.90} & \textbf{0.83} & \textbf{0.74} \\
    \end{tabular}
    }
}
\vspace{-10pt}
\end{table*}

\begin{table}[h]\small
\centering
\caption{\textbf{Generalized performance and efficiency comparison.} Models are trained on 40 Fanta can trajectories. Evaluation across Fanta, Coke, spray, and hand targets demonstrates the robust zero-shot capability and real-time inference speed ($f$) of \rgmps{}. See details in \cref{sec:sota}.}
\label{table: grasp}
\vspace{-10pt}
{
\resizebox{1\columnwidth}{!}{
\setlength\tabcolsep{3pt}
\renewcommand\arraystretch{1.05}
\begin{tabular}{r||cccc|cc}

\hline\thickhline
\rowcolor{mygray}
Methods& Fanta$\uparrow$   &  Coke $\uparrow$ & Spray $\uparrow$ & Hand $\uparrow$ & Avg $\uparrow$ & $f$(Hz) $\uparrow$  \\ \hline\hline
{ManiSkill2-1st} \cite{gao2023two} &  0.70 & 0.60 &0.63 & 0.62 & 0.64 & 15.2\\ 
{Octo} \cite{team2024octo}&  0.65 & 0.55 &0.58 & 0.62 & 0.60 & 13.4\\
{OpenVLA} \cite{kim2024openvla}&  0.68 & 0.58 &0.61 & 0.60 & 0.62& 3.6\\
{RDT-1b} \cite{liu2024rdt}&  0.70 & 0.61 &0.60 & 0.62 & 0.64& 6.7\\
{DP} \cite{chi2023diffusion}&  0.75 & 0.65 &0.68 & 0.72 & 0.70& 1.01 \\
{Dex-VLA} \cite{wen2025dexvla}&  0.87 & 0.66 &0.71 & 0.84 & 0.77& 55.3 \\
\hline
Our RGMP \cite{FCCL_CVPR22}  &  \textbf{0.98} & \underline{0.78} &\underline{0.81} & \underline{0.90} & \underline{0.87} & \textbf{78.6}\\
Our \rgmps{}   &  \underline{0.96} & \textbf{0.83} &\textbf{0.85} & \textbf{0.93} & \textbf{0.89} & \underline{75.2}

\end{tabular}
}}
\vspace{-0.4cm}
\end{table}

\begin{table}[h]\small
\centering
\caption{\textbf{Ablation study of \wqkv{} and GMM.} Quantitative evaluations on tissue passing and squashed Coke manipulation tasks validate the effectiveness of the proposed components. See details in \cref{sec:ablation}.}
\label{table: abla1}
\vspace{-10pt}
{
\resizebox{1\columnwidth}{!}{
\setlength\tabcolsep{3pt}
\renewcommand\arraystretch{1.05}
\begin{tabular}{r|c||ccc|ccc}

\hline\thickhline
\rowcolor{mygray}
 & & \multicolumn{3}{c|}{Tissue} & \multicolumn{3}{c}{Squashed Coke} \\
\rowcolor{mygray}
\multirow{-2}{*}{Methods} &\multirow{-2}{*}{GMM} & $Acc_s\uparrow$ & $Acc_t\uparrow$ & $Acc\uparrow$ & $Acc_s\uparrow$ & $Acc_t\uparrow$ & $Acc\uparrow$ \\
\hline\hline
 & -- & 0.85 & 0.58 & 0.50 & 0.80 & 0.61 & 0.49 \\
\multirow{-2}{*}{DP \cite{chi2023diffusion}} & $\checkmark$ & 0.80 & 0.68 & 0.56 & 0.85 & 0.65 & 0.55 \\
\hline
& -- & 0.80 & 0.69 & 0.55 & 0.85 & 0.71 & 0.60 \\
\multirow{-2}{*}{Our \wqkv{}} & $\checkmark$ & \textbf{0.85} & \textbf{0.71} & \textbf{0.60} & \textbf{0.90} & \textbf{0.77} & \textbf{0.69}

\end{tabular}
}}
\vspace{-0.4cm}
\end{table}

\noindent \textbf{Long-horizon Manipulation Analysis}. 
We summarize the comparative results of long-horizon manipulation tasks in Figure \ref{fig:long}. The figure clearly depicts that under complex physical constraints, standard vision-language models like Octo and OpenVLA present significantly suboptimal performance in these three tasks. This demonstrates the difficulty of handling complex stochastic physical properties without explicit geometric guidance. Our \rgmps{} framework consistently outperforms all other counterparts on different scenarios. For example, in the Pouring Water scenario, \rgmps{} outperforms the best baseline Diffusion Policy by approximately 16\%. This metric is calculated through the successful transfer of liquid into target receptacles with randomized spatial configurations and various container geometries. We further conduct experiments within the towel folding scenario involving highly deformable objects, and our method achieves a superior success rate of 0.86 compared to 0.68 for Dex-VLA. This validates the robust generalization capability of the proposed framework in aligning fabric edges and maintaining stable folded geometry within high-degree-of-freedom environments. Similarly, in the Bin Picking task characterized by complex occlusions, \rgmps{} attains the highest success rate of 0.83 and effectively avoids unintended collisions. Relative to the preliminary RGMP conference version \cite{FCCL_CVPR22}, \rgmps{} incorporates dense spiking features to enrich spatiotemporal representations. This mechanism significantly enhances the robustness of action execution across the temporal dimension for long-horizon tasks and consequently yields consistently superior performance over the previous version RGMP.

\begin{table}[!t]\small
\centering
\caption{\textbf{Component contribution analysis.} Quantitative evaluations across grasping generalization tasks validate the efficacy of RoPE, Spatial Mixing Block (SMB), Channel Mixing Block (CMB), and Guided Self-attention Module (GSM).  Please see details in \cref{sec:ablation}.}
\label{table: abla2}
\vspace{-10pt}
{
\resizebox{1\columnwidth}{!}{
\setlength\tabcolsep{3pt}
\renewcommand\arraystretch{1.05}
\begin{tabular}{cccc|cccc}
\hline\thickhline
\rowcolor{mygray}
RoPE & SMB & CMB & GSM & Fanta $\uparrow$ & Coke $\uparrow$ & Spray $\uparrow$ & Hand $\uparrow$ \\
\hline\hline
-- & $\checkmark$ & $\checkmark$ & $\checkmark$ &0.88 & 0.67 & 0.73 & 0.79 \\
$\checkmark$ & -- & $\checkmark$ & $\checkmark$ &0.85 & 0.76 & 0.76 & 0.84 \\
$\checkmark$ & $\checkmark$ & -- &$\checkmark$ & 0.92 & 0.63 & 0.66 & 0.75 \\
$\checkmark$ & $\checkmark$ & $\checkmark$ &-- & 0.86 & 0.61 & 0.62 & 0.73 \\
$\checkmark$ & $\checkmark$ & $\checkmark$ &$\checkmark$ & \textbf{0.96} & \textbf{0.83} & \textbf{0.85} & \textbf{0.93}

\end{tabular}
}}
\vspace{-10pt}
\end{table}

\begin{figure}[!t]
  \includegraphics[width=\columnwidth]{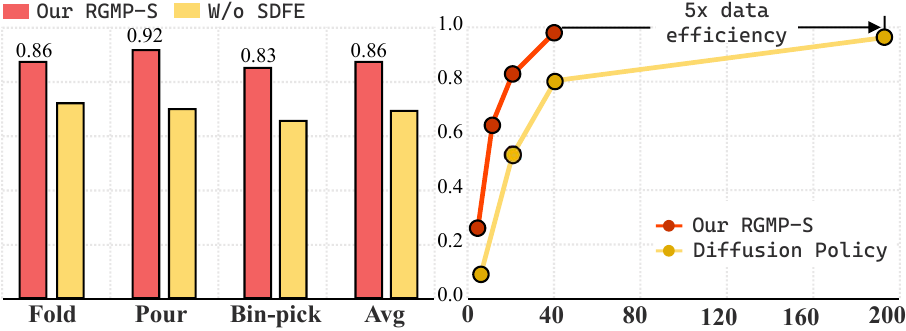}
  \vspace{-0.8cm}
\caption{\textbf{Ablation study and data efficiency analysis.} The left panel validates the Spiking Dense Feature Extraction (\spike{}) module across long-horizon tasks. The right panel illustrates that \rgmps{} achieves comparable performance to the Diffusion Policy baseline with $5\times$ fewer training trajectories on grasping Fanta task. See details in \cref{sec:ablation}.}
\label{fig:abla}
\vspace{-0.4cm}
\end{figure}

\begin{table*}[t]\small
\centering
\caption{\textbf{Quantitative comparison on the ManiSkill2 benchmark.} Comparison of average success rates (in \%, mean $\pm$ standard deviation) of our \rgmps{} against state-of-the-art baselines across five diverse manipulation tasks. All results are averaged over multiple random seeds. See \cref{sec:sota}.}
\label{tab:mani}
\vspace{-10pt}
{
    \resizebox{0.9\linewidth}{!}{
\begin{tabular}{r||ccccc|c}
\hline\thickhline
\rowcolor{mygray}
Methods& Push Chair$\uparrow$   & Move
Bucket $\uparrow$ & Plug
Charger $\uparrow$ &
Cabinet Door $\uparrow$ &
Cabinet Drawer $\uparrow$ & Average Score $\uparrow$ \\ \hline\hline
RDT-1b \cite{liu2024rdt}&  3.97 $\pm$ 1.35 & 1.46 $\pm$ 0.34 & 1.75 $\pm$ 1.21 & 13.44 $\pm$ 3.86 & 11.62 $\pm$ 3.12 & 6.44 $\pm$ 1.98
\\ 
Dex-VLA \cite{wen2025dexvla}&  2.36 $\pm$ 1.33 & 3.47 $\pm$ 1.36 & 1.62 $\pm$ 1.35 & 18.36 $\pm$ 2.93 & 8.49 $\pm$ 2.48 & 6.86 $\pm$ 1.89
\\
{ManiSkill2-1st \cite{gao2023two}} &  8.63 $\pm$ 2.36 & 4.23 $\pm$ 1.39  & 3.41 $\pm$ 1.42 & 24.68 $\pm$ 3.22 & 12.48 $\pm$ 3.28 & 10.69 $\pm$ 2.33
\\  
{Octo \cite{team2024octo}} &  5.35 $\pm$ 1.24 & 2.46 $\pm$ 1.26 & 3.29 $\pm$ 1.48  & 20.37 $\pm$ 3.98 & 10.35 $\pm$ 1.48 & 8.36 $\pm$ 1.89
\\     
{OpenVLA \cite{kim2024openvla}} &  6.33 $\pm$ 1.48 & 4.49 $\pm$ 1.63 & 4.28 $\pm$ 3.21 & 24.29 $\pm$ 4.25 & 15.47 $\pm$ 2.33 & 10.97 $\pm$ 2.58
\\    
{Diffusion Policy \cite{chi2023diffusion}} &  4.39 $\pm$ 1.22 & 6.48 $\pm$ 2.39 &6.44 $\pm$ 2.36 & 15.32 $\pm$ 1.48 & 17.38 $\pm$ 2.68 & 10.01 $\pm$ 2.03
\\
\hline
{Our RGMP \cite{FCCL_CVPR22}} & 14.32 $\pm$ 2.32 & 8.46 $\pm$ 1.43 & 7.44 $\pm$ 1.28 & 26.48 $\pm$ \textbf{4.35} & 20.46 $\pm$ 3.45 & 15.43 $\pm$ 2.57\\
\textbf{Our \rgmps{}}  &  \textbf{16.47 $\pm$ 2.34} & \textbf{9.64 $\pm$ 1.46} &\textbf{8.45 $\pm$ 2.23} & \textbf{26.48} $\pm$ 4.26 & \textbf{20.16 $\pm$ 4.18} & \textbf{16.24 $\pm$ 2.89}
    \end{tabular}
    }
}
\vspace{-10pt}
\end{table*}

\begin{figure*}[!ht]
  \includegraphics[width=\textwidth]{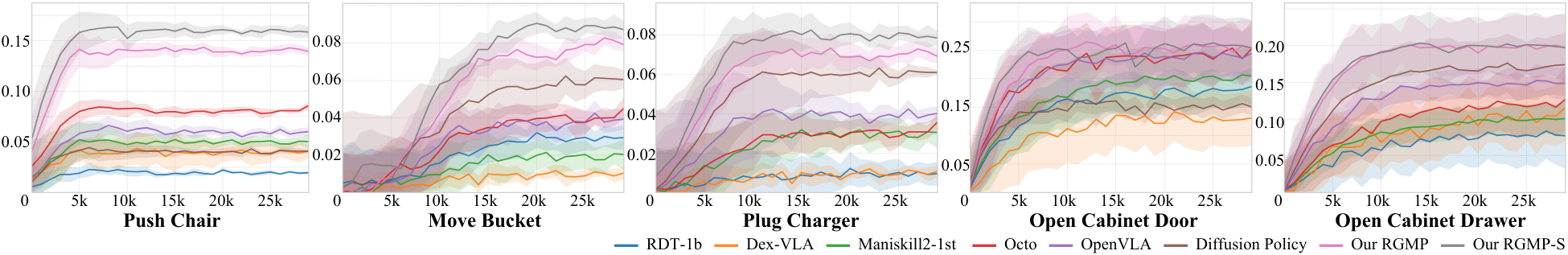}
  \vspace{-0.8cm}
\caption{\textbf{Learning curves on ManiSkill2 benchmark.} The X-axis represents training steps, and the Y-axis denotes the success rate. Shaded areas correspond to the standard deviation. Our \rgmps{} variants achieve stable convergence and superior performance across all tasks. Refer to \cref{sec:sota}.}
\label{fig:curves}
\vspace{-0.4cm}
\end{figure*}

\begin{figure}[!t]
  \includegraphics[width=\columnwidth]{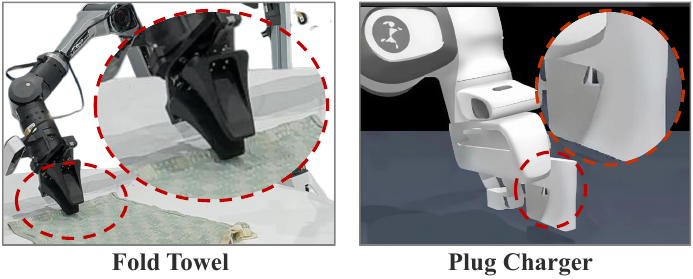}
  \vspace{-0.8cm}
\caption{\textbf{Failure cases.} Our method and current state-of-the-art models struggle in scenarios requiring fine-grained manipulation. Refer to \cref{sec:ablation}.}
\label{fig:failed}
\vspace{-0.4cm}
\end{figure}

\noindent \textbf{Simulation Benchmark Analysis}. We present a comprehensive evaluation on the ManiSkill2 benchmark, summarized in \cref{tab:mani} and visualized via the radar chart in \cref{fig:mani} and learning curves in \cref{fig:curves}. As observed, Transformer-based and Diffusion Policy methods exhibit suboptimal performance when subjected to stochastic physical constraints. This degradation highlights the inherent challenge of adapting to randomized friction parameters without explicit geometric guidance. In contrast, our \rgmps{} demonstrates superior robustness, consistently outperforming state-of-the-art baselines across all evaluated scenarios. Specifically, in the \textit{Open Cabinet Door} task involving articulated objects, \rgmps{} achieves a success rate of 26.48\%, surpassing the ManiSkill2-1st baseline (24.68\%) and Diffusion Policy (15.32\%). This performance advantage is attributed to our method's ability to handle stochastic dynamics while actuating target components to at least 90\% of their maximum range. Furthermore, in the bimanual \textit{PushChair} task, \rgmps{} achieves a score of 16.47\%, nearly doubling the performance of the strongest external baseline (ManiSkill2-1st at 8.63\%). This significant margin validates the efficacy of our approach in preventing object tipping and ensuring precise static positioning within 15 cm of the target destination.

\subsection{Diagnostic Experiments}
\label{sec:ablation}
\noindent \textbf{Ablation study}. To rigorously evaluate the individual contributions of the proposed components within our \rgmps{} framework, we conducted a series of ablation studies across diverse manipulation tasks. First, we investigated the efficacy of the Spiking Dense Feature Extraction (SDFE) module in long-horizon tasks. As illustrated in the left panel of Fig. \ref{fig:abla}, removing the SDFE module leads to a significant performance degradation across all tested tasks (Fold, Pour, Bin-pick). For instance, in the "Pour" task, the success rate drops from 0.92 to approximately 0.65 without SDFE. This confirms that the SDFE module is crucial for capturing temporal dependencies and fine-grained spatial features necessary for complex, multi-stage manipulation. Second, we analyzed the impact of the \wqkv{} module and the GMM component. Table \ref{table: abla1} presents quantitative results on tissue passing and squashed Coke manipulation tasks. The baseline Diffusion Policy achieves a combined accuracy ($Acc$) of 0.50 and 0.49 on the Tissue and Squashed Coke tasks, respectively. Incorporating GMM into the baseline improves these scores to 0.56 and 0.55. However, our proposed \wqkv{} method, even without GMM, outperforms the baseline, achieving accuracies of 0.55 and 0.60. The combination of \wqkv{} and GMM yields the highest performance, reaching an accuracy of 0.60 on Tissue and 0.69 on Squashed Coke. This demonstrates that \wqkv{} effectively enhances feature representation, and its synergy with GMM further enhances the robustness of the policy against object deformability and state uncertainty. Third, we performed a component contribution analysis to validate the efficacy of Rotary Positional Embeddings (RoPE), the Spatial Mixing Block (SMB), the Channel Mixing Block (CMB), and the Guided Self-attention Module (GSM) in grasping generalization tasks. As detailed in Table \ref{table: abla2}, the full model achieves the highest success rates across all objects (Fanta: 0.96, Coke: 0.83, Spray: 0.85, Hand: 0.93). Removing any single component results in a noticeable performance drop. For example, omitting the GSM reduces the success rate on the "Hand" object from 0.93 to 0.73, highlighting its importance in guiding attention to relevant geometric features. Similarly, the absence of SMB leads to a significant decrease in performance on the "Fanta" object (0.96 to 0.85), underscoring the necessity of spatial mixing for robust grasping.

\noindent \textbf{Data Efficiency Analysis.} We further evaluated the data efficiency of \rgmps{} compared to the Diffusion Policy on the Fanta grasping task. The right panel of Fig. \ref{fig:abla} illustrates the dependency of the success rate on the quantity of training trajectories. Our method demonstrates superior data efficiency, achieving a high success rate with significantly fewer demonstrations. Specifically, \rgmps{} reaches a performance level comparable to the converged Diffusion Policy (trained on $\sim$200 trajectories) using only $\sim$40 trajectories. This 5$\times$ improvement in data efficiency is attributed to the inductive biases introduced by our \wqkv{}, which allow for effective learning from sparse data.

\noindent \textbf{Failure Case Analysis and Limitations.} To ensure a comprehensive and rigorous evaluation, we investigate the boundary conditions where the proposed \rgmps{} and other state-of-the-art baselines encounter difficulties. As visualized in \cref{fig:failed}, performance degradation is observed across all evaluated methods in scenarios demanding extreme perceptual precision. Specifically, in the \textit{Fold Towel} task, when the deformable fabric lies completely flush against the table surface, the available vertical clearance for gripper insertion diminishes to the millimeter level. This insufficient manipulation margin allows for negligible error tolerance, frequently causing unintended collisions with the tabletop or failed grasping attempts due to precise z-axis control limits. Similarly, in the \textit{Plug Charger} task, the success rate drops significantly when the insertion alignment tolerance is stricter than 5 mm. Under such high-precision constraints, self-occlusion of the robotic gripper hinders the fine-grained visual feedback required for exact mating. It is worth noting that these failure modes reflect the intrinsic complexity of the physical interaction rather than specific algorithmic flaws. Empirical observations indicate that these scenarios are non-trivial even for human operators, where the one-shot success rate via teleoperation is also limited. These findings highlight the open challenges in handling high-fidelity contact dynamics and suggest future directions for integrating tactile feedback or active vision.

\section{Conclusion}
In this paper, we present \rgmps{}, a novel and robust framework for generalizable robotic manipulation. \rgmps{} exploits both geometric-prior knowledge and spatiotemporal feature learning by constructing the Long-horizon Geometric-prior Skill Selector (\gss{}) and the Recursive Adaptive Spiking Network (\wqkv{}), effectively overcoming the semantic-geometric gap and acquiring generalizable dexterity in dynamic environments. Furthermore, for alleviating the data scarcity problem, we formulate an Adaptive Spike Neuron to modulate feature retention and integrate a Gaussian Mixture Model to refine action generation, enabling the policy to learn from limited demonstrations while preserving precision. Moreover, we maintain the supervision of geometric consistency to provide strong spatial constraints, boosting both long-horizon planning and fine-grained interaction performance. We experimentally demonstrate that \rgmps{} performs favorably against many existing state-of-the-art methods across ten diverse manipulation tasks, including physical interactive scenarios and standard simulation benchmarks. We hope this work paves for future research on efficient and generalizable robotic learning as it is fully reproducible and includes:

\begin{itemize}
\item \rgmps{}, a strong baseline that outperforms existing approaches while maintaining \textbf{high data efficiency} and \textbf{robust generalization} across unseen scenarios.
\item A comprehensive benchmark comparison of the state-of-the-art on \textbf{multiple real-world} and simulation tasks.
\end{itemize}

\ifCLASSOPTIONcaptionsoff
  \newpage
\fi


%
{\small
\bibliographystyle{IEEEtran}

\bibliography{egbib}
}



\vfill


\end{document}